\documentclass[12pt]{article}

\usepackage[left=1in, right=1in, top=1in, bottom=1in]{geometry}

\usepackage{titlesec}
\usepackage{setspace}
\usepackage[utf8]{inputenc}
\usepackage{csquotes}
\usepackage{xcolor}
\usepackage{lscape}
\usepackage{float}
\usepackage{enumitem}
\usepackage{caption}
\usepackage{subcaption}
\usepackage{url}
\usepackage{amsmath,amsthm,bm,mathrsfs}
\usepackage{amssymb}
\usepackage{amsfonts}
\usepackage[mathcal]{euscript}

\usepackage{graphicx}
\usepackage{booktabs}
\usepackage{tabularx}

\usepackage{natbib}

\usepackage{tikz}
\usetikzlibrary{arrows.meta}
\usetikzlibrary{shapes.geometric}

\newcommand{\XORmat}[1]{\mathrm{XOR}\text{-}#1}

\newcommand{\Supp}{\mathsf{Supp}}

\newcommand{\XORop}{\mathsf{XOR}}

\numberwithin{equation}{section}

\title{\sffamily\bfseries\Large Multiple Criteria Decision Making Under XORness}

\author{
Amin Hocine\\[-9pt]
\small Excelia Business School, CERIIM, 37200 Tours, France\\[-9pt]
\textcolor{blue}{\small \texttt{hocinea@excelia-group.com}}
}
\date{}

\titleformat{\section}
  {\sffamily\bfseries\Large\centering}  
  {}                                    
  {0pt}                                 
  {\S\thesection\\}                     

\titleformat{\subsection}
  {\large\bfseries}                     
  {\thesubsection}{0.5em}{}             

\usepackage{amsmath}
\usepackage{amssymb}

\usepackage{fancyhdr}
\usepackage[most]{tcolorbox}
\usepackage{hyperref}

\newcommand{\citationbox}{%
  \begin{tcolorbox}[
    colback=black,
    colframe=black,
    coltext=white,
    boxrule=0.5pt,
    arc=2pt,
    left=4pt,
    right=4pt,
    top=1pt,
    bottom=1pt,
    width=\textwidth
  ]
    \scriptsize
    \textbf{Please cite this paper as:} Hocine, A. (2026). Multiple criteria decision making
    under XORness. \textit{Omega}, Article 103690. \href{https://doi.org/10.1016/j.omega.2026.103690}
    {\textcolor{white}{https://doi.org/10.1016/j.omega.2026.103690}}
  \end{tcolorbox}%
}

\fancypagestyle{plain}{%
  \fancyhf{}
  \fancyfoot[C]{\citationbox}

}

\begin{document}

\setstretch{1.5}

\maketitle

\begin{spacing}{1.1}
\begin{abstract} 
\emph{How should we make decisions under XORness?}\\
XORness is a class of 
uncertainty arising from indeterminacy, exclusivity, and hesitancy. 
We develop a multiple criteria decision making framework 
that preserves XORness throughout the decision process. We define 
XOR numbers and prove their fundamental arithmetic and algebraic properties. 
Building on these results, we extend the deterministic decision matrix to 
an XOR decision matrix and introduce a compatible normalization procedure. 
Further, we propose an additive XOR aggregation operator, which preserves indeterminacy, exclusivity, and hesitancy inherent in evaluations. 
In addition, we introduce an axiomatic XORness measure that quantifies the degree of 
uncertainty associated with global performance scores and develop 
a performance--XORness index to rank alternatives. We prove 
that this index induces a total preorder over the set of 
alternatives and identify the conditions for preserving dominance. The applicability of the developed framework is demonstrated through the risk assessment of cyberthreat scenarios in energy systems, 
and recommendations are derived using the XOR--weighted sum method.
\end{abstract}

\par
\end{spacing}

\begin{spacing}{1.1}
\begin{flushright}
\textit{Decision, as we mean the word, is creative and is able to be so through the freedom which uncertainty gives for the creation of unpredictable hypotheses. 
}\\
\textcolor{blue}{--- George Shackle (1969: p. 6)}
\end{flushright}
\end{spacing}

\section{Introduction}
\noindent 
The choice of a preference aggregation model in multiple criteria decision making\footnote{This paradigm is also called multiple criteria decision analysis or multiple criteria decision aid, reflecting different conceptions (\textcolor{blue}{Wallenius et al., 2008}; \textcolor{blue}{Figueira et al., 2013}).} (MCDM) plays a central role in shaping final recommendations (\textcolor{blue}{Roy, 1996}). The weighted sum method\footnote{This method is also called simple additive weighting (\textcolor{blue}{Tzeng \& Huang, 2011}).} (WSM) remains one of the most widely adopted methods for additive aggregation in MCDM owing to its intuitive structure and computational tractability.
\textcolor{blue}{Belton and Stewart (2002)} highlight these advantages as follows:

\begin{spacing}{1.1}
\begin{center}
\begin{minipage}{.8\linewidth}
Our reason for emphasizing additive aggregation is that it is the form most easily explained to and understood by decision makers from a wide variety of backgrounds, while not placing any substantially greater restrictions on the preference structures than more complicated aggregation formulae. (p.~86)
\end{minipage}
\end{center}
\end{spacing}
We first introduce the fundamentals of the deterministic WSM. Let
$\mathcal{A} := \{A_i \mid i = 1, 2, \ldots, m\}$ be a finite set of potential actions (or alternatives) evaluated with 
respect to a finite set of attributes (or criteria)
$\mathcal{C} := \{C_j \mid j = 1, 2, \ldots, n\}$. These quantitative or qualitative criteria are typically associated with 
objectives such as maximization (e.g., profit and efficiency) or 
minimization (e.g., cost and risk). More precisely, for each \( j \in \{1,2,\dots,n\} \), we define a criterion \( C_j \) as a real-valued function $C_j : \mathcal{A} \to \mathbb{R}$, where the value $C_j(A_i) := x_{ij}$ represents the performance (or evaluation) of alternative $A_i$ under the criterion $C_j$ (\textcolor{blue}{Greco et al., 2016}). Because criteria typically involve heterogeneous scales or measurement 
units, the evaluations~$x_{ij}$ are normalized to
$r_{ij} \in [0,1]$ using a suitable technique (\textcolor{blue}{Hwang \& Yoon, 1981}). To account for the relative importance of each criterion~$C_j$, assign a weight vector 
$\mathbf{w} := (w_1, w_2, \ldots, w_n)^\top \in \mathbb{R}^n$ to each criterion, such that $w_j>0$ for all \( j \in \{1,2,\dots,n\} \) and $\sum_{j=1}^{n} w_j = 1$. These weights may be determined through
various preference modeling analyses, 
including tradeoff 
(\textcolor{blue}{Dyer \& Sarin, 1979}) and pairwise comparisons  (\textcolor{blue}{Saaty, 1980}). Alternatively, in the absence of sufficient 
reasons, they may be assumed equal. Assuming additive-preference independence among criteria (\textcolor{blue}{Keeney \& Raiffa, 1976}), the overall performance of each alternative is obtained by aggregating the weighted normalized evaluations.
In particular, for every \( A_i \in \mathcal{A} \), the global performance score is defined as $S(A_i) := \sum_{j=1}^n w_j r_{ij}.$ 
This score induces a preference structure in the set of alternatives. Thus, for any pair \( A_i, A_{i'} \in \mathcal{A} \), $A_i \succsim A_{i'}$ if and only if $S(A_i) \geq S(A_{i'}),$ implying that $A_i$ is at
least as preferred as~$A_{i'}$. The associated strict preference and indifference relations are given by $A_i \succ A_{i'}$ if and only if $S(A_i) > S(A_{i'})$ and $A_i \sim A_{i'}$ if and only if $S(A_i) = S(A_{i'})$. Finally, the most preferred alternative is defined as $A^* := \operatorname*{arg\,max}_{A_i \in \mathcal{A}} S(A_i)$. The properties of such preference relations and conditions
under which they admit numerical representations have been
extensively studied in decision theory and measurement theory
(\textcolor{blue}{Debreu, 1954}; \textcolor{blue}{Fishburn, 1970};
\textcolor {blue}{Krantz et al., 1971}). Additional theoretical foundations have been provided by \textcolor{blue}{Fishburn (1978)}, \textcolor{blue}{Vincke (1992)}, \textcolor{blue}{Bouyssou and Pirlot (2008)}, and \textcolor{blue}{Dyer (2016)}, and a comprehensive historical perspective has been provided by \textcolor{blue}{K\"{o}ksalan et al.\ (2011)}.

\bigskip

Given its straightforward and intuitive structure, the WSM has been implicitly or explicitly integrated into numerous MCDM approaches, including goal programming (\textcolor{blue}{Charnes \& Cooper, 1957}), multiattribute utility theory (MAUT; \textcolor{blue}{Keeney \& Raiffa, 1976}), analytic hierarchy process (AHP) (\textcolor{blue}{Saaty, 1980}), simple multiattribute rating technique (\textcolor{blue}{Edwards, 1977}), utilité additive (\textcolor{blue}{Jacquet-Lagreze \& Siskos, 1982}), measuring attractiveness by a categorical-based evaluation technique (\textcolor{blue}{Bana e Costa \& Vansnick, 1994}), stochastic multicriteria acceptability analysis (SMAA; \textcolor{blue}{Lahdelma et al., 1998}), and robust ordinal regression (\textcolor{blue}{Corrente et al., 2013}). Thus, performance of the WSM has been evaluated under different settings. For example, \textcolor{blue}{Triantaphyllou and Lin (1996)} investigated the sensitivity of the WSM to variations in criteria weights and alternative performance values. Similarly, \textcolor{blue}{Stewart (1995)} demonstrated through simulations that preference rankings obtained from MAUT remain stable when approximated by an additive aggregation of marginal utilities. More recently, \textcolor{blue}{Dias (2024)} derived closed-form formulas for the mean and standard deviation of SMAA that avoid the need for Monte Carlo approximation.

\bigskip

To handle uncertainty in the WSM, various approaches have been proposed to address specific types (or sources) of uncertainty, such as randomness, fuzziness, and rounding and measurement errors\footnotemark\footnotetext{Philosophically, randomness is commonly linked to probability and chance. Epistemic randomness reflects ignorance or limited predictability, whereas ontic randomness reflects an objective feature of reality, as often debated in quantum mechanics. Fuzziness is mainly associated with vagueness, especially regarding borderline cases and the Sorites paradox. Interval-valued uncertainty typically underlies incomplete evidence (\textcolor{blue}{Klir, 1991}).}. These approaches can be broadly categorized into probabilistic (P-WSM), fuzzy (F-WSM), and interval (I-WSM) approaches. P-WSM addresses randomness by assuming probability distributions for evaluations. It belongs to the broader framework of MAUT, wherein alternatives are ranked according to their expected utilities under the axioms of utility theory (\textcolor{blue}{von Neumann \& Morgenstern, 1947}; \textcolor{blue}{Savage, 1954}). A comprehensive overview of this approach is provided by \textcolor{blue}{Stewart and Durbach (2016)}. Meanwhile, F-WSM has been developed to address the vagueness inherent in human judgment, particularly when evaluations are provided by linguistic expressions. In this approach, uncertainty is formalized using fuzzy sets (\textcolor{blue}{Zadeh, 1965}) to represent the imprecision associated with evaluations or weights. Generally, triangular or trapezoidal membership functions are used for formalizing uncertainty (\textcolor{blue}{Dubois \& Prade, 1980}; \textcolor{blue}{Zimmermann, 2011}). \textcolor{blue}{Baas and Kwakernaak (1977)} were likely the first to study a multiple criteria decision problem under fuzziness conditions. Subsequently, several improvements and extensions, such as aggregation, preference modeling, and fuzzy weights, were proposed to address specific issues (see, e.g., \textcolor{blue}{Chen \& Hwang (1992)}, \textcolor{blue}{Modarres \& Sadi-Nezhad (2005)}, \textcolor{blue}{Wang (2015)}, and references cited therein). Finally, I-WSM aims to address decision making problems characterized by rounding and measurement errors. Representative I-WSM approaches formalize uncertainty by assuming upper and lower bounds (i.e., interval real numbers) for evaluations and weights. Early conceptualizations by \textcolor{blue}{Moore (1979)} laid the foundation for interval arithmetic in decision analysis, which was subsequently applied to MCDM (see, e.g., \textcolor{blue}{Ahn (2017), Fern\'{a}ndez et al. (2020)}, and references cited therein). \textcolor{blue}{Hosseinzadeh et al. (2023)} have presented an overview of advances in I-WSM.

\bigskip

An analysis of these approaches provides several notable insights. First, each WSM approach has been developed to address a distinct type of uncertainty, reflecting the various sources and forms of uncertainty. Second, formalizing uncertainty through probability distributions, fuzzy membership functions, and upper and lower bounds of evaluations and weights is a core aspect of these WSM approaches. Third, considering the varying assumptions regarding the uncertainty type, preferences of decision makers, and available information, direct comparisons between these WSM approaches are often infeasible and, in some cases, conceptually inappropriate. This noncomparability results from the fact that each approach is developed based on distinct axioms, assumptions, formalisms, and data requirements. Thus, comparative analyses should not aim to identify a universally superior framework but to identify the strengths and limitations of each framework within specific contexts. Fourth, the selection of an appropriate WSM approach should be guided by the nature of uncertainty inherent in the concerned decision problem. For instance, F-WSM is well suited for linguistic uncertainty, whereas I-WSM is appropriate when dealing with uncertainty arising from rounding errors in measurement. Thus, applications should consider the uncertainty type involved. MCDM approaches based on the rough set theory, Dempster--Shafer theory, and robust analysis also address specific sources of uncertainty (see, e.g., \textcolor{blue}{Greco et al., 2001}; \textcolor{blue}{Beynon et al., 2000}; \textcolor{blue}{Aissi \& Roy, 2010}, and references cited therein). Finally, integrated approaches, such as the combination of F-WSM with P-WSM, are promising for modeling hybrid uncertainty. Nevertheless, regardless of using a specific or integrated approach, the decision-problem characteristics must be strictly aligned with the method’s underlying assumptions. Failure to achieve this alignment can lead to large analytical distortion, as properly described by Klir in \textit{Facets of Systems Science} (\textcolor{blue}{Klir, 1991}):
\begin{spacing}{1.1}
\begin{center}
\begin{minipage}{0.8\linewidth}
Each mathematical theory is derived from some specific assumptions (axioms) and, consequently, the use of any methodological tool based upon the theory is restricted to problems that conform to these assumptions. If a problem does not conform to them and an applied mathematician trained in the methodology still wants to use it, he has to adjust (reformulate) the problem to make it fit the assumptions. This means, however, that a different problem is now solved. The problem adjustment is often not stated explicitly and, as a consequence, an impression is created that the original problem was solved while, in fact, it was not. (p. 87)
\end{minipage}
\end{center}
\end{spacing}

\bigskip

To avoid this critical issue, the concept of XOR numbers was recently proposed to model uncertainty arising from indeterminacy, exclusivity, and hesitancy, which are collectively called as XORness (\textcolor{blue}{Hocine \& Kouaissah, 2020}). An XOR number is represented by a finite nonempty set of distinct candidate values that are mutually exclusive and jointly exhaustive, where exactly one candidate is, or will become, the realized value, but which candidate is realized remains indeterminate. Conceptually, XORness emerges from the state of indeterminacy when a decision maker must select one alternative (e.g., an act, an evaluation, a scenario, a parameter, or a weight) from a well-defined set of decision alternatives. However, the decision maker cannot identify and choose the true item owing to either an epistemic limitation, wherein knowledge about these alternatives is incomplete, or an ontological condition, wherein the decision system lacks fixed or determinate features. In such cases, the decision maker must employ the exclusive disjunction (exclusive OR; XOR) as a form of alternation to handle indeterminacy. Consequently, a state of psychological hesitancy is established as multiple alternatives compete for cognitive priority, with none being sufficiently dominant to be selected. Instances of XORness are common in everyday reasoning and complex domains. For example, one might say, ``It was either Adam or John, but I could not determine who.'' In medical contexts, a doctor may note, ``The symptoms point either to a viral infection or an autoimmune disorder, but we do not know which one.'' Similar expressions occur in economic projections, wherein multiple exclusive outcomes remain indeterminate under a specific situation. For example, an economist might say, ``The gross domestic product growth is projected to be 3\%, or it may weaken to 2.7\% in the next quarter.''

\bigskip

To formalize this type of uncertainty, XOR numbers have been incorporated into MCDM. In fact, several MCDM methods have been extended to operate with XOR numbers, including XOR--AHP (\textcolor{blue}{Hocine \& Kouaissah, 2020}), XOR--analytic network process (XOR--ANP) (\textcolor{blue}{Hocine et al., 2023}), XOR--data envelopment analysis (XOR--DEA) (\textcolor{blue}{Kouaissah \& Hocine, 2022}), XOR--best worst method (XOR--BWM) (\textcolor{blue}{Kouaissah \& Hocine, 2023}), and group XOR--AHP (\textcolor{blue}{Liu et al., 2023}; \textcolor{blue}{Chen et al., 2025}). Collectively, these methods provide a systematic foundation for integrating the concept of XOR numbers into hierarchical, network, group, and efficiency-based decision models. In XOR--AHP, the deterministic pairwise comparison matrix is extended into an XOR comparison matrix to enable decision makers to express preferences using XOR numbers defined on Saaty's scale. A mathematical programming model derives crisp priority vectors from XOR comparison matrices. To account for different attitudes toward uncertainty, three variants of XOR comparison matrices are proposed, namely, optimistic, pessimistic, and neutral matrices. Furthermore, a consistency index measures the coherence of XOR judgments. The XOR-ANP model extends XOR--AHP to network structures incorporating feedback and interdependence between criteria.  Similarly, XOR--BWM integrates XOR comparison matrices into the BWM structure to generate crisp weights. More recent studies have investigated the use of XOR numbers in group decision making (\textcolor{blue}{Liu et al., 2023}; \textcolor{blue}{Chen et al., 2025}). These studies have focused on consistency and its improvement in XOR comparison matrices. Beyond preference modeling, XOR--DEA incorporates XOR numbers into efficiency analysis, allowing decision making units to be evaluated using XOR-based inputs and outputs while producing crisp efficiency scores. 

A central feature of existing studies is that they operate within an \textbf{\texttt{[XOR-valued input $\rightarrow$ deterministic-valued output]}} decision system. Specifically, XOR numbers are first approximated and mapped to deterministic
real numbers. Then, the final results (e.g., priority weights or efficiency
scores) are generated as crisp deterministic values. Given this modeling paradigm, a natural question follows: \textit{How can the WSM be extended to operate within a
decision system that preserves XORness throughout evaluation,
decision matrix construction, normalization, weighted normalization, and
aggregation?}

Addressing this question requires moving beyond existing developments toward a
more general decision approach. To this end, we investigate an
\textbf{\texttt{[XOR-valued input} $\rightarrow$ \texttt{XOR-valued output]}} decision system. In this
framework, every stage of the decision process, including preference
evaluation, decision matrix construction, normalization, weighted normalization,
and aggregation, is formulated explicitly in terms of XOR numbers. Hence,
XORness is preserved throughout decision making. The main
contributions of this study are summarized below.

\begin{itemize}
\item \textbf{Arithmetic operations of XOR numbers}.
We define the arithmetic operations of XOR numbers and prove 
their fundamental algebraic properties, including closure, commutativity, 
and associativity. Theoretically, we establish XOR numbers 
as a coherent mathematical structure within uncertainty theory. From an 
information-theoretic perspective, the proposed algebra ensures that the 
information encoded in indeterminacy, exclusivity, and hesitancy is 
preserved and manipulated without distortion during computation. 
Practically, because most MCDM methods rely on aggregation, normalization, 
and weighting, this contribution enables the deterministic 
MCDM approach to be systematically extended to handle XORness.

\item \textbf{XOR-based evaluation model}.
We introduce an XOR-based evaluation model 
$C_j : \mathcal{A} \rightarrow \mathbb{X}\text{OR}(\mathbb{R})$ that 
assigns an XOR number to each alternative. This model aligns with uncertainty 
theory by recognizing that precise deterministic representations 
(i.e., $C_j : \mathcal{A} \rightarrow \mathbb{R}$) are often inadequate 
in complex decision environments. Accordingly, 
the resulting XOR decision matrix preserves XORness throughout evaluation and avoids information loss at early analytical stages. Practically, the model allows decision makers to express preferences in a cognitively realistic manner, accommodating indeterminacy, exclusivity, and hesitancy that typically arise in real evaluations.

\item \textbf{Normalization of the XOR decision matrix}. We propose a normalization technique fully compatible with the 
algebraic structure of XOR numbers. Theoretically, it 
is consistent with the principle of information-preserving transformations central 
to uncertainty and information theory. More specifically, the technique preserves the structure of XORness 
through elementwise operations on XOR numbers. By mapping 
all XOR evaluations into dimensionless values within 
$\mathbb{X}\text{OR}([0,1])$, the technique enables meaningful and 
comparable evaluations across heterogeneous criteria.

\item \textbf{XOR aggregation operator}. 
We introduce an additive XOR aggregation operator that synthesizes local 
XOR-performance evaluations into a global XOR-performance score and 
prove that it satisfies the boundary, commutativity, and monotonicity 
properties required of an aggregation operator. Theoretically, this 
operator naturally extends the deterministic 
weighted sum operator and defines a new class of uncertain 
aggregation operators. From an information-theoretic perspective, the 
operator preserves XORness embedded in the XOR decision matrix 
throughout the aggregation process. From a practical standpointFrom a practical standpoint, extending the WSM to render it applicable to 
XOR numbers ensures computational tractability and ease of implementation 
in real decision-support contexts.

\item \textbf{XORness measure axiomatization}. We propose an axiomatic XORness measure that quantifies the degree of 
indeterminacy, exclusivity, and hesitancy embedded in the global 
XOR-performance score of every alternative and prove that the 
proposed measure satisfies a set of axioms (i.e., degeneracy, 
symmetry, monotonicity under support inclusion, homogeneity, and 
neutrality). From an information-theoretic perspective, the 
measure represents an information descriptor that captures 
uncertainty in XOR evaluations. Practically, it provides 
discriminative decision support in cases where alternatives exhibit 
overlapping performance scores, enabling decision makers to move beyond 
purely dominance-based ranking procedures.

\item \textbf{Performance--XORness ranking index}. We derive a performance--XORness index ($\mathsf{PXness}_\gamma$) that 
simultaneously accounts for aggregated performance and the degree of 
XORness to rank alternatives. We prove that the proposed index induces 
a total preorder over the set of alternatives and identify the 
conditions under which the ranking preserves dominance. In particular, the 
parameter $\gamma \in [0,1]$ allows to incorporate 
the attitude toward XORness into the decision process.

\item \textbf{Cybersecurity risk assessment}. We apply the proposed framework to a hypothetical cybersecurity risk assessment problem involving five cyberthreat scenarios in the energy sector: ransomware attacks, coordinated cyberattacks, malware injection, insider attacks, and the exploitation of Internet of Things (IoT) devices. This application illustrates the applicability of the proposed framework.
\end{itemize}

The remainder of this paper is organized as follows. Section~2 discusses the concept of XORness. Section~3 presents the definitions related to XOR numbers and their arithmetic operations. Section~4 details the XOR--WSM. Section~5 introduces the XOR aggregation operator. Section~6 presents the XORness measure and ranking procedure. Section~7 illustrates the applicability of the XOR--WSM through a case study. Finally, Section 8 concludes the paper.

\section{XORness}
\noindent
To illustrate the concepts of \textit{XORness}
and \textit{XOR numbers}, consider the following example. Suppose a panel of
experts is asked to evaluate the risks to the sensitive data
infrastructure of a company on a nine-point Likert scale, with each integer corresponding
to a distinct threat level encompassing the least to the most impactful threats. For
instance, threats such as malware, phishing, and ransomware may be assigned
risk levels 4, 5, and 6, respectively. Given the scarcity of information on risk levels and the imperfect nature
of risk-level assignment, along with the structural complexity of cybersecurity,
precisely assigning specific risks levels to
cyberthreats may be challenging. Hence, a set of potential scenarios can be considered, each
representing a possible threat level. These scenarios are typically mutually
exclusive. However, one such scenario might lead to the following conclusion:

\begin{center}
    ``This type of cybersecurity risk has level 4, 5, or 6.''
\end{center}

\noindent
The word \textit{or} in this expression is far from incidental. In semantics,
exclusive disjunction is a form of alternation commonly employed to handle uncertainty, especially when responding to questions. As noted by
\textcolor{blue}{Lindley (1962)}, such usage reflects the way individuals convey
indeterminacy or a lack of knowledge in their choices. The use of exclusive
disjunction in the aforementioned expression is crucial because it carries functional
information about the nature and type of uncertainty. In this 
example, removing this linguistic device would result in a loss
of functional information. For a comprehensive discussion about the philosophical
significance of disjunction, the reader is referred to the work by \textcolor{blue}{Jennings (1994)}. Now, one may ask why we use the word ``or''\footnote{Russell pointed out in
\textit{An Inquiry into Meaning and Truth} that ``or'' can be used as inclusive or exclusive or. Inclusive or implies that either
of the two given options can be true or both are true simultaneously. For
example, if I say, ``I will have either coffee or tea,'' I mean that I am
willing to have either coffee or tea or both if I choose. In contrast, exclusive or, which is considered herein, indicates that only one of the two given
options can be true but not both. For example, if I say, ``You can have either
coffee or tea,'' using exclusive or, I mean that you may choose one of the two
options but not both \textcolor{blue}{(Russell, 1962)}.} to express our
judgment? As \textcolor{blue}{Russell (1962)} noted,

\begin{spacing}{1.1}
\begin{center}
    \begin{minipage}{0.8\linewidth}
        Hesitation may be observed in animals, but in them (one supposes), it
        does not find verbal expression. Human beings, seeking to express it,
        have invented the word `or'. (p.~210)
    \end{minipage}
\end{center}
\end{spacing}

\noindent
According to the \textit{Oxford English Dictionary}, ``hesitation'' is defined
as the act of pausing or delaying decision making or action due to irresolution and refers to a condition of doubt in relation to action
(\textcolor{blue}{``hesitation,'' 2024}). Etymologically, hesitation is derived
from the Latin word \textit{haesitationem}, which represents ``irresolution'' or
``uncertainty'' (\textcolor{blue}{Harper, 2024}).

\bigskip

In logic, exclusive disjunction is represented by the XOR logical operator.
This operator returns \textit{true} if and only if the operands
differ; that is, one operand is true and the other is false. A proposition
has a truth value, with \textit{true} denoted by the binary digit 1 and
\textit{false} denoted by the binary digit 0. The set of these two binary values
is called the Boolean domain. Many propositions in our daily life, such as ``$2 < 3$,'' have straightforward truth values. However, when the truth is
indeterminate, a fundamental philosophical question arises concerning the nature of
truth and indeterminacy. According to the \textit{Oxford English Dictionary},
indeterminacy refers to something that is difficult to identify exactly
(\textcolor{blue}{``indeterminacy,'' 2024}). In philosophy, the nature of indeterminacy remains debated. One view argues that indeterminacy
has an epistemological dimension, reflecting our limitations in obtaining
definitive knowledge about certain states or properties due to inherent
uncertainties, such as those observed in complex systems or in the interpretation
of language (\textcolor{blue}{Quine, 1960}). Another perspective
suggests that indeterminacy has an ontological dimension, proposing that some
aspects of reality may lack a fixed, determinate structure, implying that the
world itself may be fundamentally indeterminate
(\textcolor{blue}{Barnes \& Williams, 2011}).

\bigskip

Returning to the cybersecurity example, because classical logic operates within
the Boolean domain, exactly one candidate risk level can be
true in practice. Nevertheless, due to indeterminacy, treating the expression ``the
cybersecurity risk has level 4, 5, or 6'' as a \textit{unified cognitive
entity} that represents the overall risk assessment allows the proper handling of
the nature of uncertainty inherent in such evaluations. This perspective
aligns with the view of indeterminacy as an epistemic limitation while remaining
open to ontological possibilities. Formally, the expression
\[
4 \ \mathsf{XOR} \ 5 \ \mathsf{XOR} \ 6
\]
\noindent
represents an \textbf{XOR number}. The notions of \textit{indeterminacy},
\textit{exclusivity}, and \textit{hesitation} collectively define a type of
uncertainty called as \textbf{XORness}.

\section{XOR Numbers}
\noindent In this section, we formalize the notion of an XOR number and establish its arithmetic operations and algebraic properties.

\medskip

\noindent \textbf{Definition 1.}
Let $k \in \mathbb{N}$, with $k \geq 1$. An XOR number is a finite set of distinct real-valued elements connected by an $\mathsf{XOR}$ relation, where \emph{exactly one} element represents the true value but its identity remains indeterminate. We express an XOR number as

\begin{equation}
X := x_1 \ \mathsf{XOR}\ x_2 \ \mathsf{XOR}\ \cdots \ \mathsf{XOR}\ x_k
\end{equation}

\noindent or equivalently as
\begin{equation}
X := \mathsf{XOR}_{h=1}^{k} x_h.
\end{equation}

\noindent The set
\begin{equation}
\mathsf{Supp}(X):=\{x_1, x_2, \ldots, x_k\} \subseteq \mathbb{R}
\end{equation}
is called the \emph{support} of $X$, and its \emph{cardinality} is defined as

\begin{equation}
\mathsf{Card}(X) := |\mathsf{Supp}(X)|=k.
\end{equation}

\noindent If $\mathsf{Card}(X) = 1$, then $X$ reduces to a singleton set and corresponds to a deterministic real number.

We also use the notation
\begin{equation}
X :=\mathsf{XOR}\{ x_1, x_2, \ldots, x_k \}
\end{equation}
\noindent interchangeably with notations (3.1) and (3.2) to denote an XOR number.

\medskip

\noindent \textbf{Remark 1.} The symbol $\mathsf{XOR}$ appearing between elements is structural and not computational. It does not refer to the conventional Boolean XOR operation applied to truth values.

\noindent \textbf{Example 1.} Let $X = 1 \ \mathsf{XOR} \ 2 \ \mathsf{XOR}\ 3$ be an XOR number with cardinality $\mathsf{Card}(X) = 3$. Among the real values $\{1,2,3\}$, exactly one is realized (true value); however, its identity remains indeterminate, as illustrated in Figure 1.

\begin{figure}[h]
\centering
\begin{tikzpicture}[>=Latex, scale=0.6]

  \def\xA{1}
  \def\xB{4.5}
  \def\xC{8}
  \def\h{2.2}   

  \pgfmathsetmacro{\cx}{(\xA+\xC)/2}
  \pgfmathsetmacro{\cy}{\h/2}
  \pgfmathsetmacro{\rad}{(\xC-\xA)/2 + 0.6}

  \draw[red, thin, dashed] (\cx, \cy) circle (\rad);
  \node[red, font=\small\bfseries, anchor=south]
    at (\cx, {\cy + \rad + 0.1}) {$X$};

  \draw[->, thin] (-0.8,0) -- (9.5,0);
  \node[right] at (9.5,0) {$\mathbb{R}$};

  \foreach \x/\lab in {\xA/1, \xB/2, \xC/3}{
    \draw[line width=2pt] (\x,0) -- (\x,\h);
    \fill (\x,0) circle (4pt);
    \node[below=6pt, font=\small] at (\x,0) {$\lab$};
  }

  \draw[thick] (2.40,-0.12) -- (2.60,0.12);
  \draw[thick] (2.60,-0.12) -- (2.80,0.12);
  \draw[thick] (5.90,-0.12) -- (6.10,0.12);
  \draw[thick] (6.10,-0.12) -- (6.30,0.12);

  \draw[->, thick, blue]
    (\xA,\h) arc[start angle=180, end angle=0, radius={(\xB-\xA)/2}]
    node[midway, fill=white, inner sep=2pt, font=\small, text=blue] {$\mathsf{XOR}$};

  \draw[->, thick, blue]
    (\xB,\h) arc[start angle=180, end angle=0, radius={(\xC-\xB)/2}]
    node[midway, fill=white, inner sep=2pt, font=\small, text=blue] {$\mathsf{XOR}$};

\end{tikzpicture}
\caption{Graphical representation of the XOR number
$X = 1 \ \XORop\ 2 \ \XORop\ 3$. The three values forming the support
set lie on the real line $\mathbb{R}$. The blue arcs connect these values
through the XOR relation to indicate that exactly one of them is the
true value, but its identity remains indeterminate. The dashed red
circle encloses the XOR number $X$, emphasizing that the three
admissible values jointly constitute a single indivisible entity. The
double-slash marks on the axis between elements indicate the
discontinuity of the support, such that $\Supp(X) = \{1,2,3\}$ is a
discrete subset of $\mathbb{R}$.}
\label{fig:xor3}
\end{figure}
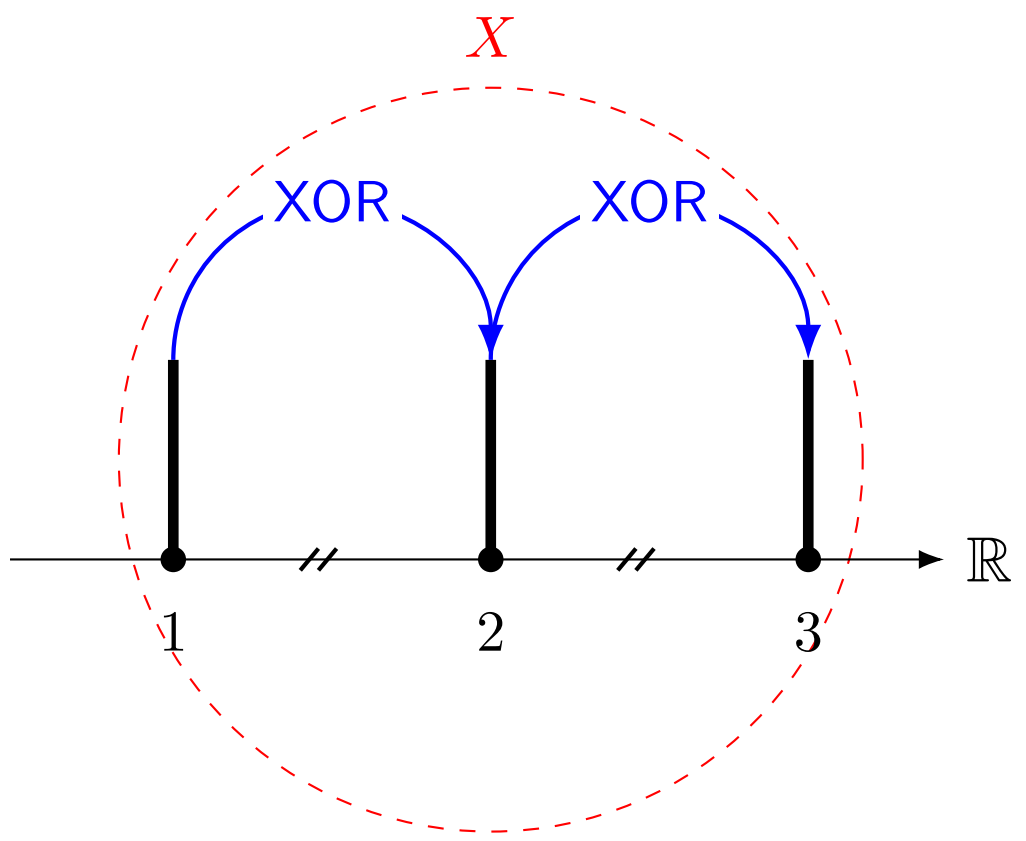

\noindent \textbf{Remark 2.} The ordering of elements in an XOR number is immaterial. Hence, the XOR relation is commutative and associative: $x_1\ \mathsf{XOR}\ x_2 = x_2\ \mathsf{XOR}\ x_1$
and $(x_1\ \mathsf{XOR}\ x_2)\  \mathsf{XOR}\ x_3 = x_1\ \mathsf{XOR}\ (x_2\ \mathsf{XOR}\ x_3).$

According to Definition 1, the set of all XOR numbers in the real line can be defined as

\begin{equation}
\mathbb{X}\text{OR}(\mathbb{R}) := \left\{ X \mid \exists k \in \mathbb{N},\, k \ge 1\!:\!
\begin{array}{l}
\exists x_1, x_2, \ldots, x_k \in \mathbb{R} :
x_p \ne x_q \ \forall p \ne q\,\\
X = \mathsf{XOR}\{x_1, x_2, \ldots, x_k\}.
\end{array}
\right\}.
\end{equation}

\medskip

\noindent \textbf{Definition 2.} 
Let $X, Y$ $\in \mathbb{X}\text{OR}(\mathbb{R})$. Then,
\begin{equation}
X = Y \iff \mathsf{Supp}(X) = \mathsf{Supp}(Y).
\end{equation}

\medskip

\noindent 
\textbf{Definition 3.} Let $X, Y$ $\in \mathbb{X}\text{OR}(\mathbb{R})$. Then,
\begin{equation}
X > Y \iff \forall x \in \mathsf{Supp}(X), \forall y \in \mathsf{Supp}(Y): x > y.
\end{equation}

\medskip

Because an XOR number represents a situation wherein exactly one element 
in its support is the true value but the identity of this element remains 
indeterminate, any arithmetic operation involving XOR 
numbers must account for all possible realizations of the operands. 
Accordingly, the support of the resulting XOR number is obtained by 
applying the given operation to all pairwise combinations of elements 
in the corresponding supports. This construction ensures the preservation of 
indeterminacy, exclusivity, and hesitancy throughout 
computation.

\medskip

\noindent \textbf{Definition 4.} Let $X, Y \in \mathbb{X}\text{OR}(\mathbb{R})$ with $\mathsf{Supp}(X)=\{x_1,x_2,\dots,x_{k_1}\}$ and $ \mathsf{Supp}(Y)=\{y_1,y_2, \dots,y_{k_2}\}.$ The \textit{sum} of $X$ and $Y$, denoted as $X \oplus Y$, is the XOR number whose support is defined as the set of all pairwise sums of the elements of $\mathsf{Supp}(X)$ and $\mathsf{Supp}(Y)$. That is, 
\begin{equation}
X \oplus Y := \mathsf{XOR} \left\{ (x_p + y_q)\ \middle|\ p =1,2,\dots,k_1,\ q =1,2,\dots,k_2 \right\}.
\end{equation}
The cardinality of the resulting XOR number is $\mathsf{Card}(X \oplus Y) \leq k_1 \cdot k_2$.

\medskip

\noindent\textbf{Example 2.} Let \(X = 1 \ \mathsf{XOR}\ 4\) and \(Y = 7 \ \mathsf{XOR}\ 8 \ \mathsf{XOR}\ 9\) be XOR numbers with $\mathsf{Supp}(X)=\{1, 4\}$ and $
\mathsf{Supp}(Y)=\{7, 8, 9\},$ where 
\(\mathsf{Card}(X) = k_1 = 2\) and \(\mathsf{Card}(Y) = k_2 = 3\). By Definition 4, the sum
\(X \oplus Y\) is obtained by taking all pairwise sums of elements in
\(\mathsf{Supp}(X)\) and \(\mathsf{Supp}(Y)\). Therefore, $X \oplus Y = 8\ \mathsf{XOR}\ 9\ \mathsf{XOR}\ 10\ \mathsf{XOR}\ 11\ \mathsf{XOR}\ 12\ \mathsf{XOR}\ 13$, and \(\mathsf{Card}(X \oplus Y) = 6 = k_1 \cdot k_2\).

\medskip

\noindent\textbf{Lemma 1.}
If $X,Y \in \mathbb{X}\mathrm{OR}(\mathbb{R})$, then $X \oplus Y \in \mathbb{X}\mathrm{OR}(\mathbb{R})$.

\noindent
\textbf{Proof.} Let $X=\mathsf{XOR}\{x_1,x_2,\dots,x_{k_1}\}$ and
$Y=\mathsf{XOR}\{y_1,y_2,\dots,y_{k_2}\}$.
By Definition 4, $X\oplus Y
=\mathsf{XOR}\{(x_p+y_q) \mid p=1,2,\dots,k_1,\ q=1,2,\dots,k_2\,\}.$ Because $\mathbb{R}$ is closed under addition, each sum $x_p+y_q$ is a real number.
Because $k_1$ and $k_2$ are finite, the support set of $X\oplus Y$ is finite. Hence, $X\oplus Y \in \mathbb{X}\mathrm{OR}(\mathbb{R})$.
\hfill$\blacksquare$

\medskip

\noindent\textbf{Lemma 2.} Let
$X,Y\in \mathbb{X}\mathrm{OR}(\mathbb{R})$, then $X\oplus Y = Y\oplus X$.

\noindent
\textbf{Proof.}
Let $X=\mathsf{XOR}\{x_1, x_2,\dots,x_{k_1}\}$ and $Y=\mathsf{XOR}\{y_1,y_2,\dots,y_{k_2}\}$.
By Definition 4, $X\oplus Y=\mathsf{XOR}\{(x_p+y_q) \mid p=1, 2, \dots,k_1,\ q=1, 2,\dots,k_2\,\}.$ Because addition in $\mathbb{R}$ is commutative, $x_p+y_q=y_q+x_p$ for all $p$ and $q$.
Therefore, $X\oplus Y=\mathsf{XOR}\{(y_q+x_p) \mid q=1,2,\dots,k_2,\ p=1,2,\dots,k_1\,\}=Y\oplus X.$
\hfill $\blacksquare$

\medskip

\noindent\textbf{Lemma 3.} If
$X,Y,Z\in \mathbb{X}\mathrm{OR}(\mathbb{R})$, then $(X\oplus Y)\oplus Z = X\oplus (Y\oplus Z).$

\noindent
\textbf{Proof.}
Let $X=\mathsf{XOR}\{x_1,x_2,\dots,x_{k_1}\}$, $Y=\mathsf{XOR}\{y_1,y_2,\dots,y_{k_2}\}$,
and $Z=\mathsf{XOR}\{z_1,z_2,\dots,z_{k_3}\}$. By Definition 4, $(X\oplus Y)\oplus Z
=\mathsf{XOR}\{\, (x_p+y_q)+z_r \mid p=1,2,\dots,k_1,\ q=1,2,\dots,k_2,\ r=1,2,\dots,k_3\,\}.$ Because addition in $\mathbb{R}$ is associative, $(x_p+y_q)+z_r=x_p+(y_q+z_r)$ for all $p$, $q$, and $r$.
Hence, $(X\oplus Y)\oplus Z
=\mathsf{XOR}\{\, x_p+(y_q+z_r) \mid p=1,2,\dots,k_1,\ q=1,2,\dots,k_2,\ r=1,2,\dots,k_3\,\}
= X\oplus (Y\oplus Z).$
\hfill $\blacksquare$

\medskip

\noindent 
\textbf{Definition 5.} 
The difference between two XOR numbers $X$ and $Y$, denoted as $X \ominus Y$, is defined as the set of all pairwise differences of the elements of $\mathsf{Supp}(X)$ and $\mathsf{Supp}(Y)$.
\begin{equation}
X \ominus Y := \mathsf{XOR} \left\{ (x_p - y_q)\ \middle|\ p =1,2,\dots,k_1,\ q =1,2,\dots,k_2 \right\}.
\end{equation}
The cardinality of the resulting XOR number is $\mathsf{Card}(X \ominus Y) \leq k_1 \cdot k_2$.

\medskip

\noindent 
\textbf{Definition 6.} The product of two XOR numbers $X$ and $Y$, denoted as $X \otimes Y$, is defined as the set of all pairwise products of the elements of $\mathsf{Supp}(X)$ and $\mathsf{Supp}(Y)$. 
\begin{equation}
X \otimes Y := \mathsf{XOR} \left\{ (x_p \times y_q)\ \middle|\ p =1,2,\dots,k_1,\ q =1,2,\dots,k_2 \right\}.
\end{equation}
The cardinality of the resulting XOR number is $\mathsf{Card}(X \otimes Y) \leq k_1 \cdot k_2$.

\medskip

\noindent \textbf{Example 3.} Let $X = 1\ \mathsf{XOR}\ 2$ and $Y = 3\ \mathsf{XOR}\ 4$ be two XOR numbers with cardinalities $\mathsf{Card}(X) = \mathsf{Card}(Y) = 2$. By Definition 6, $\mathsf{Supp}(X \otimes Y)=\{(1 \times 3), (1 \times 4), (2 \times 3), (2 \times 4)\}.$ The resulting XOR number is $X \otimes Y = 3\ \mathsf{XOR}\ 4\ \mathsf{XOR}\ 6\ \mathsf{XOR}\ 8,$ with $\mathsf{Card}(X \otimes Y) = 4.$

\medskip

\noindent\textbf{Lemma 4.}
If $X,Y \in \mathbb{X}\mathrm{OR}(\mathbb{R})$, then $X \otimes Y \in \mathbb{X}\mathrm{OR}(\mathbb{R})$.

\noindent
\textbf{Proof.}
Let $X = \mathsf{XOR}\{x_1, x_2, \dots, x_{k_1}\}$ and 
$Y = \mathsf{XOR}\{y_1, y_2, \dots, y_{k_2}\}$.
By Definition 6, $X \otimes Y
= \mathsf{XOR}\{(x_p \times y_q) \mid p=1,2,\dots,k_1,\ q=1,2,\dots,k_2\}.$ Because $\mathbb{R}$ is closed under multiplication, each product $x_p \times y_q$ is a real number.
Because there are finite products, $X\otimes Y$ has a finite support with real elements.
Hence, $X \otimes Y \in \mathbb{X}\mathrm{OR}(\mathbb{R})$.
\hfill$\blacksquare$

\medskip

\noindent \textbf{Lemma 5.}
If $X, Y \in \mathbb{X}\text{OR}(\mathbb{R})$, then $X \otimes Y = Y \otimes X.$

\noindent
\textbf{Proof.}
Let $X = \mathsf{XOR}\{x_1, x_2, \dots, x_{k_1}\}$ and
$Y = \mathsf{XOR}\{y_1, y_2, \dots, y_{k_2}\}$. By Definition 6, $X \otimes Y
=\mathsf{XOR}\{(x_p \times y_q) \mid p=1,2,\ldots,k_1,\ q=1,2,\ldots,k_2\,\}.$ Because multiplication in $\mathbb{R}$ is commutative, then
$x_p \times y_q = y_q \times x_p$ for all $p$ and $q$. Therefore, $X \otimes Y
=\mathsf{XOR}\{(y_q \times x_p) \mid q=1,2,\ldots,k_2,\ p=1,2,\ldots,k_1\,\}
= Y \otimes X.$
\hfill $\blacksquare$

\medskip

\noindent \textbf{Lemma 6.}
If 
$X, Y, Z \in \mathbb{X}\text{OR}(\mathbb{R})$, then $(X \otimes Y) \otimes Z = X \otimes (Y \otimes Z).$

\noindent
\textbf{Proof.}
Let $X =\mathsf{XOR}\{x_1, x_2, \dots, x_{k_1}\}, 
Y = \mathsf{XOR}\{y_1, y_2, \dots, y_{k_2}\}, and 
Z = \mathsf{XOR}\{z_1, z_2, \dots, z_{k_3}\}$. By Definition 6, $(X \otimes Y) \otimes Z
=\mathsf{XOR}\{\, (x_p \times y_q) \times z_r \mid
p=1,2,\ldots,k_1,\ q=1,2,\ldots,k_2,\ r=1,2,\ldots,k_3\,\}.$ Because multiplication in $\mathbb{R}$ is associative, we have $(x_p \times y_q) \times z_r = x_p \times (y_q \times z_r)$ for all $p$, $q$, and $r$. Hence, $(X \otimes Y) \otimes Z
=\mathsf{XOR}\{\, x_p \times (y_q \times z_r) \mid
p=1,2,\ldots,k_1,\ q=1,2,\ldots,k_2,\ r=1,2,\ldots,k_3\,\}
= X \otimes (Y \otimes Z)$.
\hfill $\blacksquare$

\medskip

\noindent
\textbf{Definition 7.} Let $Y$ be an XOR number such that $y_q \neq 0$ for all $q = 1,2,\ldots,k_2$. The quotient of $X$ and $Y$, denoted as $X \oslash Y$, is the XOR number defined as
\begin{equation}
X \oslash Y :=  \mathsf{XOR}\left\{ \frac{x_p}{y_q} \;\middle|\; p = 1,2,\ldots,k_1,\; q = 1,2,\ldots,k_2 \right\}.
\end{equation}
The cardinality of the resulting XOR number is $\mathsf{Card}(X \oslash Y) \leq k_1 \cdot k_2$.

Because all the aforementioned definitions share the same structure, we can unify them as follows.

\medskip

\noindent \textbf{Definition 8.}
\noindent Let $\ast \in \{ +, -, \times, \div \}$ be a binary operation on $\mathbb{R}$, with $\div$ corresponding to $\oslash$.
For $X = \mathsf{XOR}_{p=1}^{k_1} x_p$ and $Y = \mathsf{XOR}_{q=1}^{k_2} y_q$, the XOR arithmetic operation on $\mathbb{X}\text{OR}(\mathbb{R})$ is defined as 
\begin{equation}
X \circledast Y := \mathop{\mathsf{XOR}}\limits_{\substack{p = 1,\ldots,k_1 \\ q = 1,\ldots,k_2}} (x_p \ast y_q).
\end{equation}
\noindent The cardinality of the resulting XOR number is $\mathsf{Card}(X \circledast Y) \leq k_1 \cdot k_2$. The operation mechanism of (3.13) is illustrated in Figure 2.

\begin{figure}[H]
\centering
\begin{tikzpicture}[>=Latex, scale=0.78]

  \def\h{1.6}
  \def\rdot{2.5pt}

  \def\yX{9.0}
  \def\yY{4.5}
  \def\yRes{0.0}

  \def\xXa{2.0}  \def\xXb{5.0}                       
  \def\xYa{8.0}  \def\xYb{10.5} \def\xYc{13.0}       
  \def\xRa{4.5}  \def\xRb{6.5}  \def\xRc{8.5}        
  \def\xRd{10.5} \def\xRe{12.5} \def\xRf{14.5}       

  \def\xmin{-0.8}
  \def\xmax{16.5}

  \newcommand{\breakmark}[2]{%
    \draw[thick] ({#1-0.10},{#2-0.10}) -- ({#1+0.10},{#2+0.10});
    \draw[thick] ({#1+0.05},{#2-0.10}) -- ({#1+0.25},{#2+0.10});
  }

  \pgfmathsetmacro{\cxX}{(\xXa+\xXb)/2}
  \pgfmathsetmacro{\cyX}{\yX + \h/2}
  \pgfmathsetmacro{\rxX}{(\xXb-\xXa)/2 + 0.7}
  \pgfmathsetmacro{\ryX}{\h/2 + 1.5}
  \draw[red, thin, dashed] (\cxX,\cyX) ellipse[x radius=\rxX, y radius=\ryX];

  \pgfmathsetmacro{\cxY}{(\xYa+\xYc)/2}
  \pgfmathsetmacro{\cyY}{\yY + \h/2}
  \pgfmathsetmacro{\rxY}{(\xYc-\xYa)/2 + 0.7}
  \pgfmathsetmacro{\ryY}{\h/2 + 1.6}
  \draw[red, thin, dashed] (\cxY,\cyY) ellipse[x radius=\rxY, y radius=\ryY];

  \pgfmathsetmacro{\cxR}{(\xRa+\xRf)/2}
  \pgfmathsetmacro{\cyR}{\yRes + \h/2}
  \pgfmathsetmacro{\rxR}{(\xRf-\xRa)/2 + 0.9}
  \pgfmathsetmacro{\ryR}{\h/2 + 1.85}
  \draw[red, thin, dashed] (\cxR,\cyR) ellipse[x radius=\rxR, y radius=\ryR];

  \draw[->, thin] (\xmin,\yX) -- (\xmax,\yX);
  \node[right] at (\xmax,\yX) {$\mathbb{R}$};
  \node[left, font=\small\bfseries, text=red] at (\xmin,\yX) {$X:$};

  \foreach \x/\lab in {\xXa/{x_{1}}, \xXb/{x_{2}}}{
    \draw[line width=1.4pt] (\x,\yX) -- (\x,\yX+\h);
    \fill (\x,\yX) circle (\rdot);
    \node[below=4pt, font=\small] at (\x,\yX) {$\lab$};
  }

  \breakmark{({(\xXa+\xXb)/2-0.15})}{\yX}

  \draw[->, thick, blue]
    (\xXa,\yX+\h) arc[start angle=180, end angle=0, radius={(\xXb-\xXa)/2}]
    node[midway, fill=white, inner sep=2pt, font=\small, text=blue] {$\XORop$};

  \draw[->, thin] (\xmin,\yY) -- (\xmax,\yY);
  \node[right] at (\xmax,\yY) {$\mathbb{R}$};
  \node[left, font=\small\bfseries, text=red] at (\xmin,\yY) {$Y:$};

  \foreach \x/\lab in {\xYa/{y_{1}}, \xYb/{y_{2}}, \xYc/{y_{3}}}{
    \draw[line width=1.4pt] (\x,\yY) -- (\x,\yY+\h);
    \fill (\x,\yY) circle (\rdot);
    \node[below=4pt, font=\small] at (\x,\yY) {$\lab$};
  }

  \breakmark{({(\xYa+\xYb)/2-0.15})}{\yY}
  \breakmark{({(\xYb+\xYc)/2-0.15})}{\yY}

  \draw[->, thick, blue]
    (\xYa,\yY+\h) arc[start angle=180, end angle=0, radius={(\xYb-\xYa)/2}]
    node[midway, fill=white, inner sep=2pt, font=\small, text=blue] {$\XORop$};
  \draw[->, thick, blue]
    (\xYb,\yY+\h) arc[start angle=180, end angle=0, radius={(\xYc-\xYb)/2}]
    node[midway, fill=white, inner sep=2pt, font=\small, text=blue] {$\XORop$};

  \draw[->, thin] (\xmin,\yRes) -- (\xmax,\yRes);
  \node[right] at (\xmax,\yRes) {$\mathbb{R}$};
  \node[left, font=\small\bfseries, text=red] at (\xmin,\yRes) {$X \circledast Y:$};

  \foreach \x/\lab in {%
      \xRa/{x_{1}\ast y_{1}}, \xRb/{x_{1}\ast y_{2}}, \xRc/{x_{1}\ast y_{3}},
      \xRd/{x_{2}\ast y_{1}}, \xRe/{x_{2}\ast y_{2}}, \xRf/{x_{2}\ast y_{3}}}{
    \draw[line width=1.4pt] (\x,\yRes) -- (\x,\yRes+\h);
    \fill (\x,\yRes) circle (\rdot);
    \node[below=4pt, font=\scriptsize] at (\x,\yRes) {$\lab$};
  }

  \breakmark{({(\xRa+\xRb)/2-0.15})}{\yRes}
  \breakmark{({(\xRb+\xRc)/2-0.15})}{\yRes}
  \breakmark{({(\xRc+\xRd)/2-0.15})}{\yRes}
  \breakmark{({(\xRd+\xRe)/2-0.15})}{\yRes}
  \breakmark{({(\xRe+\xRf)/2-0.15})}{\yRes}

  \foreach \xa/\xb in {\xRa/\xRb, \xRb/\xRc, \xRc/\xRd, \xRd/\xRe, \xRe/\xRf}{
    \draw[->, thick, blue]
      (\xa,\yRes+\h) arc[start angle=180, end angle=0, radius={(\xb-\xa)/2}]
      node[midway, fill=white, inner sep=1pt, font=\scriptsize, text=blue] {$\XORop$};
  }

  \draw[dashed, gray] (\xXa,\yX) -- (\xRa,\yRes+\h);
  \draw[dashed, gray] (\xYa,\yY) -- (\xRa,\yRes+\h);
  \draw[dashed, gray] (\xXa,\yX) -- (\xRb,\yRes+\h);
  \draw[dashed, gray] (\xYb,\yY) -- (\xRb,\yRes+\h);
  \draw[dashed, gray] (\xXa,\yX) -- (\xRc,\yRes+\h);
  \draw[dashed, gray] (\xYc,\yY) -- (\xRc,\yRes+\h);
  \draw[dashed, gray] (\xXb,\yX) -- (\xRd,\yRes+\h);
  \draw[dashed, gray] (\xYa,\yY) -- (\xRd,\yRes+\h);
  \draw[dashed, gray] (\xXb,\yX) -- (\xRe,\yRes+\h);
  \draw[dashed, gray] (\xYb,\yY) -- (\xRe,\yRes+\h);
  \draw[dashed, gray] (\xXb,\yX) -- (\xRf,\yRes+\h);
  \draw[dashed, gray] (\xYc,\yY) -- (\xRf,\yRes+\h);

\end{tikzpicture}
\caption{Graphical representation of the unified XOR arithmetic operation
$X \circledast Y$ expressed in Definition~8, where
$\ast \in \{+,-,\times,\div\}$. The first two rows show the supports of
$X = x_{1}\,\XORop\,x_{2}$ and $Y = y_{1}\,\XORop\,y_{2}\,\XORop\,y_{3}$ on
the real line. The
last row shows the support of $X \circledast Y$ obtained by applying the
operation $\ast$ to every pair $(x_{p},y_{q})$. The dashed gray connectors
trace the $k_{1}\cdot k_{2}=6$ pairwise combinations. Each XOR number is
enclosed in a dashed red ellipse, indicating that its admissible values
jointly constitute a single indivisible entity. The double-slash marks
between elements on each axis indicate the discontinuity of the supports.}
\label{fig:xor_unified}
\end{figure}
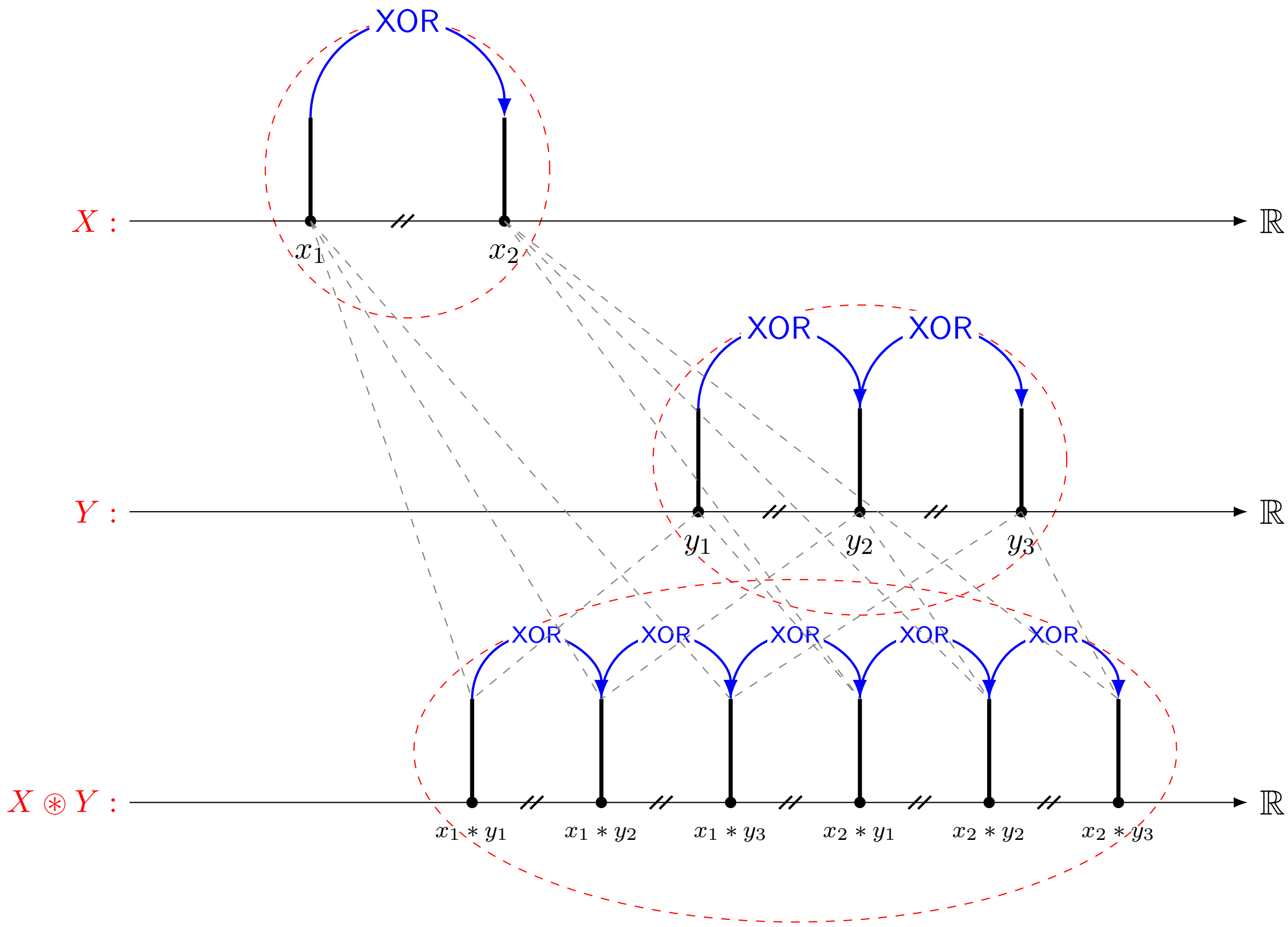

\noindent \textbf{Remark 3.}
The result of any arithmetic operation between XOR numbers is reduced to the canonical form of an XOR number by retaining only distinct elements.

\medskip

\noindent \textbf{Remark 4.} Definition 8 permits arithmetic operations between XOR numbers whose support sets may have the same or different cardinality.

\medskip

Because an XOR number is represented by a finite support set, several conventional algebraic operations can be extended to this structure, including scalar multiplication and scalar translation.

\medskip

\noindent \textbf{Definition 9.}
Let 
$\alpha \in \mathbb{R}$ and $X$ be an XOR number with 
$\mathsf{Supp}(X) = \{x_1, x_2, \ldots, x_k\}$. Scalar 
multiplication $\alpha \odot X$ is defined as
\begin{equation}
\alpha \odot X := 
\mathsf{XOR}\{(\alpha \cdot x_p) \mid p = 1, 2,\ldots, k\}.
\end{equation}
The cardinality of the resulting XOR number is 
$\mathsf{Card}(\alpha \odot X) \leq k$.

\medskip

\noindent \textbf{Example 4.} Let $X = 1\ \mathsf{XOR}\ 3$ and $\alpha = 2$. By Definition 9, 
$\alpha \odot X = (2 \cdot 1)\ \mathsf{XOR}\ (2 \cdot 3) = 2\ \mathsf{XOR}\ 6,$ with $\mathsf{Card}(\alpha \odot X) = 2.$

\medskip

\noindent \textbf{Definition 10.} Let 
$\beta \in \mathbb{R}$ and $X$ be an XOR number with 
$\mathsf{Supp}(X) = \{x_1, x_2, \ldots, x_k\}$. The translation of 
$X$ by $\beta$, denoted as $\beta \oplus X$, is defined as
\begin{equation}
\beta \oplus X := 
\mathsf{XOR}\{(\beta + x_p) \mid p = 1, 2, \ldots, k\}.
\end{equation}
The cardinality of the resulting XOR number is 
$\mathsf{Card}(\beta \oplus X) = k$.

\medskip

\noindent \textbf{Example 5.} Let $X = 1\ \mathsf{XOR}\ 3$ and $\beta = 5$. By Definition 10,
$\beta \oplus X = (5 + 1)\ \mathsf{XOR}\ (5 + 3) = 6\ \mathsf{XOR}\ 8,$ with $\mathsf{Card}(\beta \oplus X) = 2.$

\section{Modeling MCDM Under XORness}
\noindent \textcolor{blue}{Roy (1996)} stated that MCDM cannot be properly achieved without considering uncertainty during evaluation. \textcolor{blue}{Belton and Stewart (2002)} categorized uncertainty in MCDM into internal uncertainty, which is associated with the structuring and analytical process of the concerned problem, and external uncertainty, which pertains to the environment and potential consequences of specific actions (\textcolor{blue}{Kahneman \& Tversky, 1982}). Generally, an evaluation model is selected based on the nature of involved criteria (\textcolor{blue}{Ben Amor et al., 2015}). For criteria characterized by randomness, probabilistic or stochastic MCDM approaches are suitable because alternatives can be evaluated through probability distributions. For criteria characterized by vagueness, fuzzy membership functions are used to represent evaluations. In cases where rounding errors or measurement inaccuracies exist, interval MCDM allows to quantify uncertainty through interval evaluations. Thus, the following question naturally arises: \textit{If the evaluation process is conducted over a set of criteria that is characterized by XORness, how can the associated MCDM problem be effectively solved?}

In response to this, we propose the XOR--WSM for MCDM. The following argument assumes that the reader is familiar with the MCDM framework; otherwise, the reader is referred to studies by \textcolor{blue}{Tzeng and Huang (2011)}, \textcolor{blue}{Ishizaka and Nemery (2013)}, \textcolor{blue}{Greco et al. (2016)}, and \textcolor{blue}{Doumpos et al. (2019)} for fundamental concepts, methodologies, applications, and materials related to MCDM. The XOR--WSM involves the following procedures.

\bigskip

\noindent
{\sffamily\bfseries Procedure 1:} \textbf{XOR decision matrix}

\noindent
First, we construct an XOR decision matrix that elicits the preferences of the decision maker under XORness. As noted previously, in deterministic WSM approaches, each criterion is a real-valued function
\begin{equation}
C_j : \mathcal{A} \rightarrow \mathbb{R} 
\end{equation}

\noindent that assigns a deterministic real number to each alternative $A_i$ under criterion $C_j$ as follows:
\begin{equation}
C_j(A_i) := x_{ij},
\end{equation}
\noindent reflecting the performance level or score of the alternative. The collection of all such evaluations yields a decision matrix $D := (x_{ij}) \in \mathbb{R}^{m \times n}$:
\begin{equation}
D :=
\begin{pmatrix}
x_{11} & x_{12} & \cdots & x_{1n} \\
x_{21} & x_{22} & \cdots & x_{2n} \\
\vdots & \vdots & \ddots & \vdots \\
x_{m1} & x_{m2} & \cdots & x_{mn}
\end{pmatrix}.
\end{equation}
The evaluation model (4.1) assumes that the performance of each alternative is well determined and can be represented by a single real number as in (4.2). However, this assumption rarely holds in practice. Evaluation is often affected by uncertainty (e.g., imprecise measurements, incomplete data, or expert disagreement) and cognitive and psychological factors that influence human judgment. As noted by \textcolor{blue}{Guitouni and Martel (1998)}, decision makers are often emotionally and psychologically engaged in evaluation, which markedly influences the outcomes. Consequently, states of hesitancy and indeterminacy can arise, making the decision matrix (4.3) an unrealistic representation of a real-world decision problem. To address this issue, we extend the mapping function (4.1) as follows: 
\begin{equation}
C_j : \mathcal{A} \to \mathbb{X}\text{OR}(\mathbb{R}).
\end{equation}
\noindent That is, the evaluation of the alternative $A_i$ with respect to the criterion $C_j$ is an XOR number:
\begin{equation}
C_j(A_i) := X_{ij} = x_{ij}^1 \ \mathsf{XOR} \ x_{ij}^2 \ \mathsf{XOR} \ \dots \ \mathsf{XOR} \ x_{ij}^{k_{ij}} = \mathsf{XOR}_{h=1}^{k_{ij}} x_{ij}^h, 
\end{equation}
\noindent which reflects the performance of the alternative under XORness, with $
\mathsf{Card}(X_{ij}) = k_{ij}$.

\medskip

For notational simplicity, we set 
$k_{ij} = k$ for all $i,j$. Hence, $X_{ij}=\mathsf{XOR}_{h=1}^{k} x_{ij}^h$. All results extend naturally to a case with heterogeneous cardinalities. The collection of such evaluations across $A_i \in \mathcal{A}$ and $C_j \in \mathcal{C}$ defines the XOR decision matrix $\XORmat{D} \in \mathbb{X}\text{OR}(\mathbb{R})^{m \times n}$:

\begin{equation}
\XORmat{D} :=
\begin{pmatrix}
\mathsf{XOR}_{h=1}^k x_{11}^h & \mathsf{XOR}_{h=1}^k x_{12}^h & \cdots & \mathsf{XOR}_{h=1}^k x_{1n}^h \\
\mathsf{XOR}_{h=1}^k x_{21}^h & \mathsf{XOR}_{h=1}^k x_{22}^h & \cdots & \mathsf{XOR}_{h=1}^k x_{2n}^h \\
\vdots & \vdots & \ddots & \vdots \\
\mathsf{XOR}_{h=1}^k x_{m1}^h & \mathsf{XOR}_{h=1}^k x_{m2}^h & \cdots & \mathsf{XOR}_{h=1}^k x_{mn}^h
\end{pmatrix}.
\end{equation}

\noindent
Assume that all criteria are oriented toward maximization.
For any pair of alternatives $A_i,A_{i'}\in\mathcal{A}$, the natural
dominance relation associated with the XOR decision matrix in (4.6)
are defined as
\begin{equation}\label{eq:dominance}
\begin{aligned}
A_i \succ_D A_{i'}
&\iff C_j(A_i)>C_j(A_{i'})
\quad \forall j\in\{1,2,\ldots,n\},\\[4pt]
A_i \sim_D A_{i'}
&\iff C_j(A_i)=C_j(A_{i'})
\quad \forall j\in\{1,2,\ldots,n\},
\end{aligned}
\end{equation}
where $\succ_D$ and $\sim_D$ denote the strict
preference and indifference, respectively
(\textcolor{blue}{Greco et al., 2016}). The relations $>$ and $=$ on
$\operatorname{XOR}(\mathbb{R})$ are defined according to
Definitions~2 and~3.

\bigskip

\noindent
{\sffamily\bfseries Procedure 2:} \textbf{Normalized XOR decision matrix}\\
To enable the comparison of alternatives across multiple conflicting criteria, the XOR decision matrix obtained in Procedure 1 is normalized. Mapping all evaluations to the range
$\mathbb{X}\text{OR}([0,1])$ is essential for three main reasons. First, the criteria in an XOR decision matrix typically operate at heterogeneous numerical scales or measurement units. Without normalization, aggregation would be dominated by criteria with larger numerical ranges. Normalization ensures that all evaluations become dimensionless and hence comparable across criteria. Second, normalization is performed elementwise per XOR number, fully preserving the internal structure of each XOR evaluation. Finally, because the WSM relies on commensurability across criteria, scalar multiplication $w_j \,\odot\, R_{ij}$ is meaningful only when all evaluations are expressed using a common range. Accordingly, to normalize the XOR decision matrix defined in (4.6), we introduce an elementwise normalization technique.

\medskip

\noindent \textbf{Definition 11.}
Let \( \mathcal{C}_{\max} \subseteq \mathcal{C} \) be the set of oriented maximization criteria and 
\( \mathcal{C}_{\min} = \mathcal{C} \setminus \mathcal{C}_{\max} \) be the subset of oriented
minimization criteria such that
\( \mathcal{C}_{\max} \cup \mathcal{C}_{\min} = \mathcal{C} \) and
\( \mathcal{C}_{\max} \cap \mathcal{C}_{\min} = \varnothing \).

For every criterion $C_j$, define 

\begin{equation}
 R_j(X_{ij})
:=
\begin{cases}
\mathsf{XOR}\left\{\dfrac{x^{1}_{ij}}{x_j^*},\dfrac{x^{2}_{ij}}{x_j^*},\dots,\dfrac{x^{k}_{ij}}{x_j^*}\right\},
& \text{if } C_j \in \mathcal{C}_{\max}, \\[3mm]
\mathsf{XOR}\left\{1 - \dfrac{x^{1}_{ij}}{x_j^*},1 - \dfrac{x^{2}_{ij}}{x_j^*},\dots,1 - \dfrac{x^{k}_{ij}}{x_j^*} \right\},
& \text{if } C_j \in \mathcal{C}_{\min},
\end{cases} 
\end{equation}

\noindent where $x_{ij}^{h}\geq0$ for all $i$, $j$, and $h$, and
$x_j^*:=\max\{x_{ij}^{h}:i=1,\ldots,m,\ h=1,\ldots,k\}>0$.

The result is the normalized XOR decision matrix 
$\XORmat{R} \in \mathbb{X}\text{OR}([0,1])^{m \times n}$, which is defined as follows:

\begin{equation}
\XORmat{R} :=
\begin{pmatrix}
\mathsf{XOR}_{h=1}^k r_{11}^h & \mathsf{XOR}_{h=1}^k r_{12}^h & \cdots & \mathsf{XOR}_{h=1}^k r_{1n}^h \\
\mathsf{XOR}_{h=1}^k r_{21}^h & \mathsf{XOR}_{h=1}^k r_{22}^h & \cdots & \mathsf{XOR}_{h=1}^k r_{2n}^h \\
\vdots & \vdots & \ddots & \vdots \\
\mathsf{XOR}_{h=1}^k r_{m1}^h & \mathsf{XOR}_{h=1}^k r_{m2}^h & \cdots & \mathsf{XOR}_{h=1}^k r_{mn}^h
\end{pmatrix}.
\end{equation}

\medskip

Different normalization techniques can be adapted and used, and the selected method can influence the results (see, e.g., \textcolor{blue}{Hwang \& Yoon, 1981}).

\newpage

\noindent
{\sffamily\bfseries Procedure 3:} \textbf{Weighted normalized XOR decision matrix}\\
\noindent Once the normalized XOR decision matrix is obtained, the relative importance of each criterion should be defined.

\medskip

\noindent 
\textbf{Definition 12.}
Let \(\mathbf{w} = (w_1,w_2, \dots, w_n)^\top \in (0,1]^n \subseteq \mathbb{R}^n\) be a weight vector such that \(\sum_{j=1}^n w_j = 1\) and $R_{ij}$ be a normalized XOR number given by $R_{ij} = \mathsf{XOR}_{h=1}^{k}r_{ij}^{h}$. The weighted normalized XOR number corresponding to \(R_{ij}\) is defined as
\begin{equation}
V_{ij} := w_{j} \odot R_{ij} = \mathsf{XOR}_{h=1}^k \left( w_{j} \cdot r_{ij}^h \right).
\end{equation}

The weighted normalized XOR matrix
$\XORmat{V} \in\mathbb{X}\text{OR}([0,1])^{m \times n}$ is
defined as
\begin{equation}
\begin{aligned}
\XORmat{V} :=
\begin{pmatrix}

\mathsf{XOR}_{h=1}^{k} v_{11}^h & \mathsf{XOR}_{h=1}^{k} v_{12}^h & \cdots & \mathsf{XOR}_{h=1}^{k} v_{1n}^h \\
\mathsf{XOR}_{h=1}^{k} v_{21}^h & \mathsf{XOR}_{h=1}^{k} v_{22}^h & \cdots & \mathsf{XOR}_{h=1}^{k} v_{2n}^h \\
\vdots & \vdots & \ddots & \vdots \\
\mathsf{XOR}_{h=1}^{k} v_{m1}^h & \mathsf{XOR}_{h=1}^{k} v_{m2}^h & \cdots & \mathsf{XOR}_{h=1}^{k} v_{mn}^h
\end{pmatrix}.
\end{aligned}
\end{equation}

\section{XOR Aggregation Operator}
\noindent
We introduce an XOR aggregation operator that synthesizes the local XOR performances into a global XOR-performance score.

\medskip

\noindent
\textbf{Definition 13.}
Let $\mathbf{w} = (w_1,w_2,\ldots,w_n)^\top \in (0,1]^n$
such that $\sum_{j=1}^{n} w_j = 1$.
A weighted XOR aggregation operator is a mapping
\begin{equation}
\mathsf{{A}_{XOR,\mathbf{w}}}:
\mathbb{X}\text{OR}([0,1])^n
\to \mathbb{X}\text{OR}([0,1])
\end{equation}
that assigns to each $n$-tuple
$(X_1,X_2,\ldots,X_n) \in \mathbb{X}\text{OR}([0,1])^n$
an aggregated XOR number and satisfies the following
properties:

\textbf{(P1) Boundary.}
Let $0 := \mathsf{XOR}\{0\}$ and
$1 := \mathsf{XOR}\{1\}$. Then,
\begin{equation}
\mathsf{{A}_{XOR,\mathbf{w}}}(0,0,\ldots,0) = 0,
\qquad
\mathsf{{A}_{XOR,\mathbf{w}}}(1,1,\ldots,1) = 1.
\end{equation}

\textbf{(P2) Symmetry.}
For any permutation $\sigma$ of $\{1,2,\ldots,n\}$,
let
$\mathbf{w}^{\sigma}
:= (w_{\sigma(1)},w_{\sigma(2)},\ldots,w_{\sigma(n)})^\top$.
Then,
\begin{equation}
\mathsf{{A}_{XOR,\mathbf{w}}}(X_1,X_2,\ldots,X_n)
=
\mathsf{{A}_{XOR,\mathbf{w}^{\sigma}}}
(X_{\sigma(1)},X_{\sigma(2)},\ldots,X_{\sigma(n)}).
\end{equation}

\textbf{(P3) Monotonicity.} For any two vectors
$(X_1,X_2,\ldots,X_n),(Y_1,Y_2,\ldots,Y_n)
\in \mathbb{X}\text{OR}([0,1])^n$,
\begin{equation}
X_j > Y_j \quad \forall j \in \{1,2,\ldots,n\}
\quad \Longrightarrow \quad
\mathsf{{A}_{XOR,\mathbf{w}}}(X_1,X_2,\ldots,X_n)
>
\mathsf{{A}_{XOR,\mathbf{w}}}(Y_1,Y_2,\ldots,Y_n).
\end{equation}

For further details on these properties, see, for example, \textcolor{blue}{Yager (1988)}.

\bigskip

\noindent
{\sffamily\bfseries Procedure 4:}
\textbf{Aggregating local XOR performances}

\noindent
After obtaining the normalized evaluations
$(R_{i1},R_{i2},\ldots,R_{in})$ and their weighted
counterparts $V_{ij} = w_j \odot R_{ij}$ for each
alternative $A_i$ through Procedures~1--3, these local
evaluations are aggregated into a global
performance score.

\medskip

\noindent
\textbf{Theorem 1.}
Let $\mathbf{w} = (w_1,w_2,\ldots,w_n)^\top \in (0,1]^n$
be a weight vector such that $\sum_{j=1}^{n} w_j = 1$.
Define
$\mathsf{{A}_{XOR,\mathbf{w}}}:
\mathbb{X}\text{OR}([0,1])^n
\to \mathbb{X}\text{OR}([0,1])$
by

\[
\mathsf{{A}_{XOR,\mathbf{w}}}(R_1,R_2,\ldots,R_n)
:= \bigoplus_{j=1}^{n}(w_j \odot R_j).
\]
For each alternative $A_i$, let
$(R_{i1},R_{i2},\ldots,R_{in})$ be its normalized
XOR local performances, with
$V_{ij} = w_j \odot R_{ij}$. Then,
\begin{equation}
\mathsf{{A}_{XOR,\mathbf{w}}}(R_{i1},R_{i2},\ldots,R_{in})
=
\bigoplus_{j=1}^{n}(w_j \odot R_{ij})
=
\bigoplus_{j=1}^{n}V_{ij}.
\end{equation}
The operator $\mathsf{{A}_{XOR,\mathbf{w}}}$ satisfies
properties (P1)--(P3).

\noindent
\textbf{Proof.}
Let $(R_1,R_2,\ldots,R_n)
\in \mathbb{X}\text{OR}([0,1])^n$.
By Definitions~4 and~9, every element of the aggregated
support has the form $\sum_{j=1}^{n} w_j r_j$, where
$r_j \in \mathrm{Supp}(R_j) \subseteq [0,1]$.
Since $w_j > 0$ and $\sum_{j=1}^{n} w_j = 1$, $0 \leq \sum_{j=1}^{n} w_j r_j \leq 1.$ The aggregated support is finite and nonempty. Thus, $\mathsf{{A}_{XOR,\mathbf{w}}}$ is well-defined with values in $\mathbb{X}\text{OR}([0,1])$. (P1) If every normalized evaluation equals
$R_j = 0 = \mathsf{XOR}\{0\}$, then, by
Definitions~4 and~9, $\mathsf{{A}_{XOR,\mathbf{w}}}(0,0,\ldots,0)
=
\mathsf{XOR}\!\left\{\sum_{j=1}^{n} w_j \cdot 0\right\}
=
\mathsf{XOR}\{0\}
=
0.$ Similarly, if every normalized evaluation equals
$R_j = 1 = \mathsf{XOR}\{1\}$, then $\mathsf{{A}_{XOR,\mathbf{w}}}(1,1,\ldots,1)
=
\mathsf{XOR}\!\left\{\sum_{j=1}^{n} w_j \cdot 1\right\}
=
\mathsf{XOR}\{1\}
=
1.$ Hence, property (P1) holds. (P2) Let $\sigma$ be any permutation of
$\{1,2,\ldots,n\}$ and let
$\mathbf{w}^{\sigma}
= (w_{\sigma(1)},w_{\sigma(2)},\ldots,w_{\sigma(n)})^\top$.
Because XOR addition is commutative (Lemma~2)
and associative (Lemma~3), $\mathsf{{A}_{XOR,\mathbf{w}^{\sigma}}}
(R_{\sigma(1)},R_{\sigma(2)},\ldots,R_{\sigma(n)})
=
\bigoplus_{j=1}^{n}
(w_{\sigma(j)} \odot R_{\sigma(j)})
=
\bigoplus_{j=1}^{n}(w_j \odot R_j)
=
\mathsf{{A}_{XOR,\mathbf{w}}}(R_1,R_2,\ldots,R_n).$ Therefore, property (P2) holds. (P3) Let
$(R_1,R_2,\ldots,R_n),(R'_1,R_2,\ldots,R'_n)
\in \mathbb{X}\text{OR}([0,1])^n$
and assume $R_j > R'_j$ for every
$j \in \{1,2,\ldots,n\}$.
By Definition~3, for every
$r_j \in \mathrm{Supp}(R_j)$ and every
$r'_j \in \mathrm{Supp}(R'_j)$, we have $r_j > r'_j$.
Because $w_j > 0$, $\sum_{j=1}^{n} w_j r_j > \sum_{j=1}^{n} w_j r'_j.$ By Definitions~4 and~9, every element of the support
of $\mathsf{{A}_{XOR,\mathbf{w}}}(R_1,R_2,\ldots,R_n)$
has the form $\sum_{j=1}^{n} w_j r_j$,
and every element of the support of
$\mathsf{{A}_{XOR,\mathbf{w}}}(R'_1,R'_2,\ldots,R'_n)$
has the form $\sum_{j=1}^{n} w_j r'_j$.
The inequality therefore holds for every pair of
elements from these two supports. Consequently,
by Definition~3, $\mathsf{{A}_{XOR,\mathbf{w}}}(R_1,R_2,\ldots,R_n)
>
\mathsf{{A}_{XOR,\mathbf{w}}}(R'_1,R_2,\ldots,R'_n).$ Thus, property (P3) holds.
\hfill $\blacksquare$

\medskip

\noindent \textbf{Remark 5.}
The order relation $>$ on $\mathbb{X}\text{OR}(\mathbb{R})$ (Definition~3) is a strict partial order. Consequently, when the supports of two global XOR-performance scores overlap, neither score is dominant. Hence, we develop a ranking procedure inducing a total preorder over the set of alternatives $\mathcal{A}$, as detailed in Procedure~5.

\section{XORness Measure}
\noindent 
In theories of uncertainty, the measure of uncertainty for a given formalization (e.g., probability or fuzzy set theory) is a function that assigns to each representation of evidence in the theory (e.g., probability distribution or membership function) a nonnegative real number (\textcolor{blue}{Klir, 2006}). This number reflects the degree of each uncertainty type, such as randomness and fuzziness. In this section, we introduce a measure of XORness.

\medskip

\noindent\textbf{Definition 14.}
An \emph{XORness measure} $\mathsf{Xness(\cdot)}$ is a mapping
\begin{equation}
\mathsf{Xness} : \mathbb{X}\text{OR}(\mathbb{R}) \to \mathbb{R}_{\ge 0}
\end{equation}
that satisfies the following axioms:

\textbf{(A1) Degeneracy.}
\begin{equation}
\mathsf{Xness}(X)=0 \iff \mathsf{Card}(X)=1.
\end{equation}

\textbf{(A2) Symmetry.}
If $\mathsf{Supp}(X)=\{x_{1},x_{2},\ldots,x_{k}\}$, then, for any permutation
$\sigma$ of $\{1,2,\ldots,k\}$,
\begin{equation}
\mathsf{Xness(XOR}\{x_{1},x_{2},\ldots,x_{k}\})
=
\mathsf{Xness(XOR}\{x_{\sigma(1)},x_{\sigma(2)},\ldots,x_{\sigma(k)}\}).
\end{equation}

\textbf{(A3) Monotonicity under support inclusion.}
For any $X,Y \in \mathbb{X}\text{OR}(\mathbb{R})$,
\begin{equation}
\mathsf{Supp}(X) \subset \mathsf{Supp}(Y)
\;\Rightarrow\;
\mathsf{Xness}(X) < \mathsf{Xness}(Y).
\end{equation}

\textbf{(A4) Homogeneity.}
For any $\alpha \in \mathbb{R}$,
\begin{equation}
\mathsf{Xness}(\alpha \odot X)=|\alpha|\,\mathsf{Xness}(X).
\end{equation}

\textbf{(A5) Neutrality.}
For any $\beta \in \mathbb{R}$,
\begin{equation}
\mathsf{Xness}(\beta \oplus X)=\mathsf{Xness}(X).
\end{equation}

XORness reflects the degrees of indeterminacy, exclusivity, and hesitancy embedded in an XOR number. Thus, for the mapping (6.1) to be a meaningful XORness measure, it must satisfy certain requirements that capture the meaning of an intuitively acceptable characterization of the concept represented by the term ``degree of XORness.'' Specifically, any meaningful XORness measure should satisfy five requirements (axioms). \textbf{Degeneracy (A1)} states that an XOR number has no XORness if and only if its cardinality equals one. In this case, the XOR number reduces to a singleton real number with no indeterminacy regarding the realized element. Consequently, neither exclusivity nor hesitation exists. Any mapping intended to quantify XORness that assigns a positive value to a singleton XOR number would therefore contradict the fundamental interpretation of XOR numbers. Regarding \textbf{symmetry (A2)}, an XOR number represents a set of mutually exclusive elements, wherein uncertainty pertains to the realized element rather than to the ordering of the elements. Accordingly, the XORness measure must be invariant under any permutation of the elements. Violation of this axiom would implicitly introduce extraneous information through ordering, which is inconsistent with the definition of XOR numbers. \textbf{Monotonicity under support inclusion (A3)} requires the XORness measure to strictly increase when the existing support is enlarged by adding distinct elements. When a new distinct element is added, the number of mutually exclusive elements increases, increasing the level of indeterminacy and hesitancy associated with the XOR number. This monotonicity axiom ensures XORness sensitivity to the cardinality of the support of the XOR number. \textbf{Homogeneity (A4)} should be satisfied under scalar multiplication. The operation $\alpha \odot X$ scales all elements of the XOR number, proportionally expanding or contracting pairwise gaps among them. Because these gaps reflect the numerical magnitude of hesitancy, XORness should scale accordingly. \textbf{Neutrality (A5)} imposes invariance under translation. The operation $\beta \oplus X$ shifts all elements of an XOR number by a constant without altering their relative configuration or exclusivity structure. Hence, the degree of XORness remains unchanged. This axiom guarantees that XORness depends on the magnitude of hesitancy among elements rather than on their absolute position on the real line.

\bigskip

Because the identity of the true value remains unknown, the decision maker is compelled to deliberate among the competing elements in an XOR number. As discussed in Section 2, such a deliberation manifests as a cognitive state wherein the decision maker cycles between mutually exclusive elements. The intensity of this process depends on the number of competing elements, which determines the scope of exclusivity, and magnitude of pairwise separations among them, which reflects the degree of hesitancy. Accordingly, the pairwise absolute differences between the elements of an XOR number intuitively quantify XORness. The following theorem formalizes this reasoning.

\medskip

\noindent\textbf{Theorem 2.}
Let $X \in \mathbb{X}\text{OR}(\mathbb{R})$ be an XOR number with
$\mathsf{Supp}(X)=\{x_{1},x_{2},\ldots,x_{k}\}$ and $k=\mathsf{Card}(X)$.
Define $\mathsf{Xness}:\mathbb{X}\text{OR}(\mathbb{R})\to\mathbb{R}_{\ge0}$ as
\begin{equation}
\mathsf{Xness}(X):=
\begin{cases}
0, & \mathsf{Card}(X)=1,\\[2mm]
\dfrac{2}{\mathsf{Card}(X)}\displaystyle\sum_{1\le p<q\le \mathsf{Card}(X)}\left|x_{p}-x_{q}\right|,
& \mathsf{Card}(X)\ge 2,
\end{cases}
\end{equation}
which satisfies axioms (A1)--(A5).

\noindent
\textbf{Proof.} Let $X \in \mathbb{X}\text{OR}(\mathbb{R})$ with
$\mathrm{Supp}(X)=\{x_{1},x_{2},\ldots,x_{k}\}$ and $k=\mathsf{Card}(X)$.

\noindent (A1) If $\mathsf{Card}(X)=1$ (i.e., $k=1$), then $
\mathsf{Xness}=0$ by (6.7).
Conversely, if $\mathsf{Card}(X)\ge2$, because the elements of $\mathrm{Supp}(X)$ are distinct,
there exists at least one pair $(p,q)$ such that $x_{p}\neq x_{q}$.
Hence, $\sum_{1\le p<q\le k}|x_{p}-x_{q}|>0,$ which implies $\mathsf{Xness}(X)>0$ by (6.7). (A2) Let $\sigma$ be any permutation of $\{1,2,\dots,k\}$ and define $X' = \mathsf{XOR}\{x_{\sigma(1)},x_{\sigma(2)},\dots,x_{\sigma(k)}\}.$ If $k=1$, then $\mathsf{Xness}(X)=\mathsf{Xness}(X')=0$ by (6.7).
If $k\ge2$, the set of unordered index pairs $\{p,q\}$ with
$1\le p<q\le k$ is invariant under reindexing. Hence, $\sum_{1\le p<q\le k}|x_{p}-x_{q}|
=
\sum_{1\le p<q\le k}\bigl|x_{\sigma(p)}-x_{\sigma(q)}\bigr|.$ Therefore, by (6.7), $\mathsf{Xness}(X)
=
\frac{2}{k}\sum_{1\le p<q\le k}|x_{p}-x_{q}|
=
\frac{2}{k}\sum_{1\le p<q\le k}\bigl|x_{\sigma(p)}-x_{\sigma(q)}\bigr|
=
\mathsf{Xness}(X').$ (A3) Let $\mathrm{Supp}(Y)=\mathrm{Supp}(X)\cup\{t\},
t\notin\mathrm{Supp}(X).$ Define $A=\sum_{1\le p<q\le k}|x_{p}-x_{q}|$ and $B=\sum_{p=1}^{k}|t-x_{p}|.$ Then, $\mathsf{Xness}(X)=\frac{2}{k}A$ and 
$\mathsf{Xness}(Y)=\frac{2}{k+1}(A+B).$ By the triangle inequality $|x_{p}-x_{q}|\le |x_{p}-t|+|t-x_{q}|,$ summing over all $1\le p<q\le k$ yields $A\le (k-1)B.$ Because $t\notin\mathrm{Supp}(X)$, we have $B>0$; thus, $A<kB$.
Consequently, $\frac{2}{k}A < \frac{2}{k+1}(A+B),$ which implies that $\mathsf{Xness}(X)<\mathsf{Xness}(Y)$. (A4) For any $\alpha\in\mathbb{R}$, $\sum_{1\le p<q\le k}|\alpha x_{p}-\alpha x_{q}|
=|\alpha|\sum_{1\le p<q\le k}|x_{p}-x_{q}|,$ which implies that $\mathsf{Xness}(\alpha\odot X)=|\alpha|\mathsf{Xness}(X)$ based on (6.7). (A5) For any $\beta\in\mathbb{R}$, $\sum_{1\le p<q\le k}|(x_{p}+\beta)-(x_{q}+\beta)|
=\sum_{1\le p<q\le k}|x_{p}-x_{q}|$; therefore, $\mathsf{Xness}(\beta \oplus X)=\mathsf{Xness}(X)$ as per (6.7).
\hfill $\blacksquare$

\medskip

\noindent \textbf{Remark 6.}
The scaling factor $\frac{2}{\mathsf{Card}(X)}$ in (6.7) ensures that the XORness measure increases with increasing dispersion and cardinality of the XOR number. Hence, adding new candidates strictly increases indeterminacy, exclusivity, and hesitancy.

\bigskip

\newpage

\noindent
{\sffamily\bfseries Procedure 5:} \textbf{Ranking under XORness}

\noindent Procedure 5 ranks the alternatives based on performance and uncertainty information. Thus, the performance level is always considered while considering the decision maker's attitude toward XORness.

\medskip

\noindent
\textbf{Definition 15.}
Let $\mathcal{A} = \{A_1, A_2, \ldots, A_m\}$ be the set of 
alternatives. For each alternative $A_i \in \mathcal{A}$, let 
$S_i$ denote the global XOR-performance 
score determined using the XOR aggregation operator~(5.5). The set of all 
global XOR-performance scores is given by
\begin{equation}
\mathcal{S}
:=
\{S_i\mid A_i\in\mathcal{A}\}
=
\{S_1,S_2,\ldots,S_m\}
\subseteq \mathbb{X}\text{OR}([0,1]),
\end{equation}

\noindent where 
$\mathsf{Supp}(S_i)$ represents the set of all possible aggregated 
performance scores of alternative $A_i$ and 
$\mathsf{Card}(S_i)$ is the number of distinct possible 
realizations resulting from aggregation.

\medskip

\noindent 
\textbf{Definition 16.} 
Let $S_i \in \mathcal{S}$ be the global XOR-performance score of 
alternative $A_i$. The arithmetic mean of the possible performance 
values of $S_i$ is defined as
\begin{equation}
\mathsf{Perf}(S_i)
:=
\frac{1}{\mathsf{Card}(S_i)}
\sum_{s \in \mathsf{Supp}(S_i)} s. 
\end{equation} 

\noindent The normalized performance is defined as 
\begin{equation}
\overline{\mathsf{Perf}}(S_i):=
\frac{\mathsf{Perf}(S_i)}{\max_{i}\mathsf{Perf}(S_i)},
\end{equation}

\noindent and the normalized XORness measure is defined as

\begin{equation}
\overline{\mathsf{Xness}}(S_i):=
\frac{\mathsf{Xness}(S_i)}{\max_{i} \mathsf{Xness}(S_i)}.
\end{equation}

\medskip

\noindent 
\textbf{Definition 17.} 
For fixed parameter $\gamma \in [0,1]$, performance--XORness index $\mathsf{PXness}_{\gamma}$ of alternative $A_i$ 
is defined as
\begin{equation}
\mathsf{PXness}_{\gamma}(A_i)
:= \gamma \, \overline{\mathsf{Perf}}(S_i)
+ (1 - \gamma)(1-\overline{\mathsf{Xness}}(S_i)).
\end{equation}

\medskip

\noindent\textbf{Proposition 1.}
Fix $\gamma \in [0,1]$ and define relation 
$\succsim_{\mathsf{PXness}_{\gamma}}$ on the set of alternatives 
$\mathcal{A}$ as 
\begin{equation}
A_i \succsim_{\mathsf{PXness}_{\gamma}} A_{i'}
\quad \Longleftrightarrow \quad
\mathsf{PXness}_\gamma(A_i) \ge \mathsf{PXness}_\gamma(A_{i'}).
\end{equation}

\noindent Then, $\succsim_{\mathsf{PXness}_{\gamma}}$ is a total preorder 
on $\mathcal{A}$.

\medskip

\noindent\textbf{Proof.}
Fix $\gamma \in [0,1]$. For any alternative $A_i$, we have
$\mathsf{PXness}_\gamma(A_i) \ge \mathsf{PXness}_\gamma(A_i)$
because relation $\ge$ on $\mathbb{R}$ is reflexive.
Hence $A_i \succsim_{\mathsf{PXness}_{\gamma}} A_i$. Let $A_i$ and 
$A_{i'}$ be any two alternatives. Because $\mathsf{PXness}_\gamma(A_i)$ and 
$\mathsf{PXness}_\gamma(A_{i'})$ are real numbers and order $\ge$ on 
$\mathbb{R}$ is total, either
$\mathsf{PXness}_\gamma(A_i) \ge \mathsf{PXness}_\gamma(A_{i'})$
or
$\mathsf{PXness}_\gamma(A_{i'}) \ge \mathsf{PXness}_\gamma(A_i)$.
Thus, either $A_i \succsim_{\mathsf{PXness}_{\gamma}} A_{i'}$ or 
$A_{i'} \succsim_{\mathsf{PXness}_{\gamma}} A_i$. Let $A_i, A_{i'}$, and $A_{i''}$ 
satisfy $A_i \succsim_{\mathsf{PXness}_{\gamma}} A_{i'}$ and 
$A_{i'} \succsim_{\mathsf{PXness}_{\gamma}} A_{i''}$.
Then, $\mathsf{PXness}_\gamma(A_i) \ge \mathsf{PXness}_\gamma(A_{i'})$
and $\mathsf{PXness}_\gamma(A_{i'}) \ge \mathsf{PXness}_\gamma(A_{i''}).$ 
Because $\ge$ on $\mathbb{R}$ is transitive, 
$\mathsf{PXness}_\gamma(A_i) \ge \mathsf{PXness}_\gamma(A_{i''})$.
Hence, $A_i \succsim_{\mathsf{PXness}_{\gamma}} A_{i''}$. Therefore, 
$\succsim_{\mathsf{PXness}_{\gamma}}$ is a total preorder on the set 
of alternatives $\mathcal{A}$. 
\hfill $\blacksquare$

\medskip

\noindent\textbf{Remark 7.}
The strict preference and indifference relations associated with
$\succsim_{\mathsf{PXness}_{\gamma}}$ are given by

\[
A_i\succ_{\mathsf{PXness}_{\gamma}}A_{i'}
\iff
\mathsf{PXness}_{\gamma}(A_i)
>
\mathsf{PXness}_{\gamma}(A_{i'}),
\]
and

\[
A_i\sim_{\mathsf{PXness}_{\gamma}}A_{i'}
\iff
\mathsf{PXness}_{\gamma}(A_i)
=
\mathsf{PXness}_{\gamma}(A_{i'}).
\]

\medskip

\noindent\textbf{Definition 18.} 
Let $S_i, S_{i'} \in \mathcal{S}$ be the global XOR-performance scores 
of alternatives $A_i$ and $A_{i'}$, respectively. Alternative $A_i$ is 
said to \emph{dominate $A_{i'}$ by $(\mathsf{Perf},\mathsf{Xness})$} if
\begin{equation}
\mathsf{Perf}(S_i) \geq \mathsf{Perf}(S_{i'}) \quad \text{and} \quad 
\mathsf{Xness}(S_i) \leq \mathsf{Xness}(S_{i'}),
\end{equation}
with at least one strict inequality.

\medskip
\noindent\textbf{Proposition 2.} 
If $A_i$ dominates $A_{i'}$ by $(\mathsf{Perf},\mathsf{Xness})$, then $\mathsf{PXness}_{\gamma}(A_i) \geq \mathsf{PXness}_{\gamma}(A_{i'})$ for every $\gamma\in[0,1].$

\noindent\textbf{Proof.} By 
Definition~18, we have 
$\mathsf{Perf}(S_i) \geq \mathsf{Perf}(S_{i'})$ and 
$\mathsf{Xness}(S_i) \leq \mathsf{Xness}(S_{i'})$, with at least one 
strict inequality. Normalization by a positive constant preserves 
both weak and strict inequalities; hence, 
$\overline{\mathsf{Perf}}(S_i) \geq \overline{\mathsf{Perf}}(S_{i'})$ 
and 
$\overline{\mathsf{Xness}}(S_i) \leq 
\overline{\mathsf{Xness}}(S_{i'})$. 
The latter implies 
$1 - \overline{\mathsf{Xness}}(S_i) \geq 
1 - \overline{\mathsf{Xness}}(S_{i'})$. Therefore,

\[
\mathsf{PXness}_{\gamma}(A_i) 
= \gamma\,\overline{\mathsf{Perf}}(S_i) + (1-\gamma)\bigl(1 - \overline{\mathsf{Xness}}(S_i)\bigr) 
\geq \gamma\,\overline{\mathsf{Perf}}(S_{i'}) + (1-\gamma)\bigl(1 - \overline{\mathsf{Xness}}(S_{i'})\bigr) 
= \mathsf{PXness}_{\gamma}(A_{i'}).
\]
\hfill $\blacksquare$

We now examine the relationship between $\gamma$ and 
the dominance relation defined in~(\ref{eq:dominance}), restricting attention to maximization criteria. Suppose alternative 
$A_i$ dominates alternative $A_{i'}$ in the sense of~(\ref{eq:dominance}), 
that is, $X_{ij} > X_{i'j}$ following Definition~3 for every 
criterion $j \in \{1,2,\ldots,n\}$. By Theorem~1~(P3), the corresponding 
global XOR-performance scores satisfy $S_i > S_{i'}$, 
such that every element of $\mathsf{Supp}(S_i)$ strictly exceeds every 
element of $\mathsf{Supp}(S_{i'})$. Hence,
\begin{equation}
\mathsf{Perf}(S_i) 
\;=\; \frac{1}{\mathsf{Card}(S_i)} \sum_{s \in \mathsf{Supp}(S_i)} s 
\;>\; \frac{1}{\mathsf{Card}(S_{i'})} \sum_{s' \in \mathsf{Supp}(S_{i'})} s' 
\;=\; \mathsf{Perf}(S_{i'}).
\end{equation}
Because normalization by the common positive constant 
$\max_i \mathsf{Perf}(S_i)$ preserves strict inequality, 
$\overline{\mathsf{Perf}}(S_i) > \overline{\mathsf{Perf}}(S_{i'})$. 
Hence, the performance ($\mathsf{Perf}(\cdot)$) component of $\mathsf{PXness}_{\gamma}$ always 
preserves the dominance principle defined in~(\ref{eq:dominance}). 
Note that the XORness component ($\mathsf{Xness}(\cdot)$) behaves differently, and the following 
theorem examines this behavior.

\medskip
\noindent\textbf{Theorem 3.} Let $A_i \succ_D A_{i'}$ in the sense of
(\ref{eq:dominance}), and let $S_i$ and $S_{i'}$ be their global
XOR-performance scores, respectively. Define

\[
\Delta_P :=
\overline{\mathsf{Perf}}(S_i)
-
\overline{\mathsf{Perf}}(S_{i'}),
\qquad
\Delta_X :=
\overline{\mathsf{Xness}}(S_i)
-
\overline{\mathsf{Xness}}(S_{i'}).
\]
Then $\Delta_P>0$, and, for every $\gamma\in[0,1]$,
\begin{equation}\label{eq:pxness-diff}
\mathsf{PXness}_{\gamma}(A_i)
-
\mathsf{PXness}_{\gamma}(A_{i'})
=
\gamma\,\Delta_P-(1-\gamma)\,\Delta_X.
\end{equation}
In particular,
\begin{itemize}
\item[(i)]
if $\Delta_X\leq0$, then
$\mathsf{PXness}_{\gamma}(A_i)
\geq\mathsf{PXness}_{\gamma}(A_{i'})$
for every $\gamma\in[0,1]$;

\item[(ii)]
if $\Delta_X>0$, then
$\mathsf{PXness}_{\gamma}(A_i)
\geq\mathsf{PXness}_{\gamma}(A_{i'})$
for every $\gamma\in[\gamma^*,1]$, where
\begin{equation}\label{eq:gamma-star}
\gamma^*
=
\frac{\Delta_X}{\Delta_P+\Delta_X}
\in(0,1).
\end{equation}
\end{itemize}

\noindent\textbf{Proof.}
By the definition of $\succ_D$,
$X_{ij}>X_{i'j}$ for every criterion $j$.
For maximization criteria, normalization divides all evaluations
under criterion $C_j$ by the same positive constant $x_j^*$.
Therefore, $R_{ij}>R_{i'j}$ for every $j$.
Theorem~1~(P3) then yields $S_i>S_{i'}$, meaning that every element
of $\mathsf{Supp}(S_i)$ strictly exceeds every element of
$\mathsf{Supp}(S_{i'})$. Consequently,
$\mathsf{Perf}(S_i)>\mathsf{Perf}(S_{i'})$.
Normalization by the common positive performance maximum
preserves this strict inequality, so $\Delta_P>0$. Identity~(\ref{eq:pxness-diff}) follows directly from
Definition~17. (i) If $\Delta_X\leq0$, then
$\gamma\,\Delta_P\geq0$ and
$-(1-\gamma)\,\Delta_X\geq0$
for every $\gamma\in[0,1]$. (ii) If $\Delta_X>0$, $\gamma\,\Delta_P-(1-\gamma)\,\Delta_X\geq0$ is equivalent to $\gamma(\Delta_P+\Delta_X)\geq\Delta_X.$ Because $\Delta_P+\Delta_X>0$, this holds if and only if
$\gamma\geq\gamma^*$, where
$\gamma^*=\Delta_X/(\Delta_P+\Delta_X)$. The inequalities $\Delta_X>0$ and $\Delta_P>0$ imply
$0<\gamma^*<1$. \hfill $\blacksquare$

\noindent\textbf{Corollary 1.}
The relation $\succsim_{\mathsf{PXness}_{\gamma}}$ is monotonic with respect to
the dominance defined in~(\ref{eq:dominance}) for every $\gamma\in[\gamma^{*},1]$
when $\Delta_X > 0$, and for every $\gamma\in[0,1]$ when $\Delta_X \leq 0$.

\medskip
\noindent\textbf{Proof.}
The result follows from Theorem~3 and the definition of
$\succsim_{\mathsf{PXness}_{\gamma}}$
in Proposition~1. \hfill $\blacksquare$

\medskip
\noindent \textbf{Definition 19.}
Let $X_{ij}$ and $X_{i'j}$ be the XOR evaluations of alternatives
$A_i$ and $A_{i'}$ under criterion $C_j$, respectively.
Alternative $A_i$ is said to
\emph{dominate $A_{i'}$ by translation}, denoted by
$A_i \succ_T A_{i'}$, if and only if there exists a vector
$(\beta_1,\beta_2,\ldots,\beta_n)\in\mathbb{R}_{>0}^{n}$ such that
\begin{equation}\label{eq:transl-dom}
X_{ij}=\beta_j\oplus X_{i'j}
\qquad
\text{for every }j\in\{1,2,\ldots,n\}.
\end{equation}

\medskip
\noindent \textbf{Proposition 3.}
If $A_i \succ_T A_{i'}$, then $\mathsf{PXness}_{\gamma}(A_i)
-
\mathsf{PXness}_{\gamma}(A_{i'})
=
\gamma\,\Delta_P \geq 0$
for every $\gamma \in [0,1]$.

\medskip
\noindent \textbf{Proof.}
Suppose $A_i \succ_T A_{i'}$.
By Definition~19, there exist constants $\beta_j>0$, such that
$X_{ij}=\beta_j\oplus X_{i'j}$ for every criterion $j$. Because all criteria are oriented toward maximization, normalization divides the evaluations under criterion $C_j$ by the same positive constant $x_j^*$. Therefore, $R_{ij}
=
\delta_j \oplus R_{i'j}$, where  $\delta_j := \frac{\beta_j}{x_j^*}>0.$ By the definition of the additive XOR aggregation operator, $\mathsf{Supp}(S_i)
=
\left\{
\sum_{j=1}^{n}w_j(r_j+\delta_j)
\;\middle|\;
r_j \in \mathsf{Supp}(R_{i'j}),\ j=1,2,\ldots,n
\right\}.$ Hence, $\mathsf{Supp}(S_i)
=
\{\,s+\beta \mid s \in \mathsf{Supp}(S_{i'})\,\}$, where $\beta=\sum_{j=1}^{n}w_j\delta_j.$ 
Since every $w_j>0$, every $\delta_j>0$, we have $\beta>0$. Thus, $S_i=\beta\oplus S_{i'}.$ Translation is injective and therefore preserves the cardinality
of the support: $\mathsf{Card}(S_i)=\mathsf{Card}(S_{i'}).$ By Definition~16, $\mathsf{Perf}(S_i)
=
\frac{1}{\mathsf{Card}(S_{i'})}
\sum_{s\in\mathsf{Supp}(S_{i'})}(s+\beta)
=
\mathsf{Perf}(S_{i'})+\beta.$ Moreover, the neutrality axiom~(A5) gives $\mathsf{Xness}(S_i)
=
\mathsf{Xness}(\beta\oplus S_{i'})
=
\mathsf{Xness}(S_{i'}).$ Normalization by the common positive denominators yields $\Delta_P
=
\frac{\beta}{\max_{\ell}\mathsf{Perf}(S_{\ell})}>0$, and thus $\Delta_X
=
\overline{\mathsf{Xness}}(S_i)
-
\overline{\mathsf{Xness}}(S_{i'})
=0.$ Using the definition of the performance--XORness index,
we obtain $\mathsf{PXness}_{\gamma}(A_i)
-
\mathsf{PXness}_{\gamma}(A_{i'})
=
\gamma\,\Delta_P-(1-\gamma)\,\Delta_X
=
\gamma\,\Delta_P.$ This difference is nonnegative for every $\gamma\in[0,1]$. \hfill $\blacksquare$

\medskip
\noindent\textbf{Corollary 2.}
The relation
$\succsim_{\mathsf{PXness}_{\gamma}}$ is monotonic with respect to
the dominance by translation defined
in~(\ref{eq:transl-dom}) for every $\gamma\in[0,1]$.

\medskip
\noindent\textbf{Proof.}
The result follows from Proposition~3 and the definition of
$\succsim_{\mathsf{PXness}_{\gamma}}$
in Proposition~1. \hfill $\blacksquare$

\medskip
\noindent\textbf{Remark 8.}
Parameter $\gamma$ controls the tradeoff between performance and
XORness. When $\gamma=1$, the ranking depends solely on normalized
performance. When $\gamma=0$, it depends solely on normalized
XORness. By Theorem~3, the dominance defined
in~(\ref{eq:dominance}) is  preserved
for every $\gamma\in[\gamma^*,1]$ when $\Delta_X>0$, and for every
$\gamma\in[0,1]$ when $\Delta_X\leq0$.
By Proposition~3, dominance by translation defined
in~(\ref{eq:transl-dom}) is
preserved for every $\gamma\in[0,1]$.

\bigskip

This modeling of preferences is consistent with long-standing 
foundations in decision theory, where preferences under uncertainty rely 
on different formalizations  
(\textcolor{blue}{Wald, 1950};   
\textcolor{blue}{Savage, 1954};  
\textcolor{blue}{Arrow, 1965}; 
\textcolor{blue}{Markowitz, 1952}). Because the global 
XOR performance is represented by a finite set of mutually exclusive 
realizations, no realization is more plausible than 
another. Without relevant evidence, all mutually exclusive 
realizations are treated as epistemically equivalent. Hence, the arithmetic mean provides a natural and unbiased 
representation of the performance of aggregated XOR evaluations. 
The choice of the arithmetic mean is consistent with the Laplace 
principle of insufficient reason 
(\textcolor{blue}{Laplace, 1995}). In this setting, the XORness 
measure is analogous to variance, regret, or other 
dispersion measures in risk analysis. It captures the degree of 
indeterminacy, exclusivity, and hesitancy associated with each global 
XOR-performance score. Accordingly, the $\mathsf{PXness}_\gamma$ index 
establishes a natural analog of mean--variance analysis, where 
$\mathsf{Perf}(\cdot)$ corresponds to the mean and 
$\mathsf{Xness}(\cdot)$ plays the role of a cardinality-sensitive dispersion penalty. The proposed approach combines performance and uncertainty information and establishes conditions under which dominance between alternatives are preserved.

\section{Cyberthreat Risk Assessment}
\noindent 
We present a hypothetical application evaluating five cyberthreat scenarios for Italy's energy system: ransomware attacks (scenario 1), coordinated cyberattacks (scenario 2), malware injection (scenario 3), insider attacks (scenario 4), and IoT exploitation (scenario 5). Detailed descriptions of these scenarios are provided in the Supplementary Materials. These scenarios are assessed using five criteria: execution likelihood, intrusion detection time, severity of grid disruption, response and mitigation time, and attribution difficulty. The criteria include both quantitative and qualitative measures. For the qualitative criteria, a nine-point Likert scale is used, where 1 represents the lowest level and 9 represents the highest level of likelihood, severity, or difficulty. All criteria are considered minimization objectives, with each criterion being assigned a priority weight to reflect its relative importance. Table~\ref{tab:xor-eval} summarizes the data used for the application.

\begin{table}[ht]
\centering
\caption{Decision matrix with XOR evaluations.}
\label{tab:xor-eval}
\resizebox{\textwidth}{!}{%
\begin{tabular}{l l l l l l}
\hline
& Execution Likelihood & Intrusion Detection Time & Severity of Grid Disruption & Response \& Mitigation Time & Attribution Difficulty \\
\hline
Direction & Min & Min & Min & Min & Min \\
Type & Qualitative & Quantitative & Qualitative & Quantitative & Qualitative \\
Weight & 0.25 & 0.10 & 0.30 & 0.20 & 0.15 \\
\hline
Scenario 1 & $2 \;\mathsf{XOR}\; 3$ & $35 \;\mathsf{XOR}\; 40$ & $3 \;\mathsf{XOR}\; 4 \;\mathsf{XOR}\; 5 \;\mathsf{XOR}\; 9$ & $12 \;\mathsf{XOR}\; 13$ & $5 \;\mathsf{XOR}\; 7$ \\
Scenario 2 & $2 \;\mathsf{XOR}\; 3 \;\mathsf{XOR}\; 4$ & $35 \;\mathsf{XOR}\; 38$ & $6 \;\mathsf{XOR}\; 7 \;\mathsf{XOR}\; 8$ & $11 \;\mathsf{XOR}\; 12$ & $3 \;\mathsf{XOR}\; 6$ \\
Scenario 3 & $3 \;\mathsf{XOR}\; 5$ & $35 \;\mathsf{XOR}\; 37$ & $7 \;\mathsf{XOR}\; 8$ & $11 \;\mathsf{XOR}\; 13$ & $6 \;\mathsf{XOR}\; 8 \;\mathsf{XOR}\; 9$ \\
Scenario 4 & $1 \;\mathsf{XOR}\; 5$ & $35 \;\mathsf{XOR}\; 37 \;\mathsf{XOR}\; 38$ & $6 \;\mathsf{XOR}\; 8$ & $11 \;\mathsf{XOR}\; 12 \;\mathsf{XOR}\; 13$ & $3 \;\mathsf{XOR}\; 5$ \\
Scenario 5 & $1 \;\mathsf{XOR}\; 2 \;\mathsf{XOR}\; 3 \;\mathsf{XOR}\; 4 \;\mathsf{XOR}\; 5$ & $34 \;\mathsf{XOR}\; 38$ & $3 \;\mathsf{XOR}\; 5$ & $11 \;\mathsf{XOR}\; 13$ & $8 \;\mathsf{XOR}\; 9$ \\
\hline
\end{tabular}%
}
\end{table}

To solve the XOR decision matrix as presented in Table~\ref{tab:xor-eval}, the number of deterministic WSM problems required for analysis grows exponentially as follows:
\begin{equation}
\text{Total deterministic WSM problems} := \prod_{i=1}^{m} \prod_{j=1}^{n} 
\mathsf{Card}(X_{ij}),
\end{equation}
where \( \mathsf{Card}(X_{ij}) \) is the cardinality of XOR local performances. \noindent Figure 3 shows the complexity of the XOR decision matrix, which makes the deterministic WSM impractical. Moreover, even if computationally feasible, the need to generate and evaluate millions of decision matrices introduces additional methodological challenges, such as matrix fusion and normalization, handling conflicting rankings, and maintaining interpretability across a vast scenario space. To overcome these computational issues, we apply the developed XOR-WSM. We compute the normalized XOR decision matrix using (4.8), obtaining the results listed in Table~\ref{tab:normalized-xor}.

\begin{figure}[H]
\centering
\includegraphics[width=0.9\textwidth]{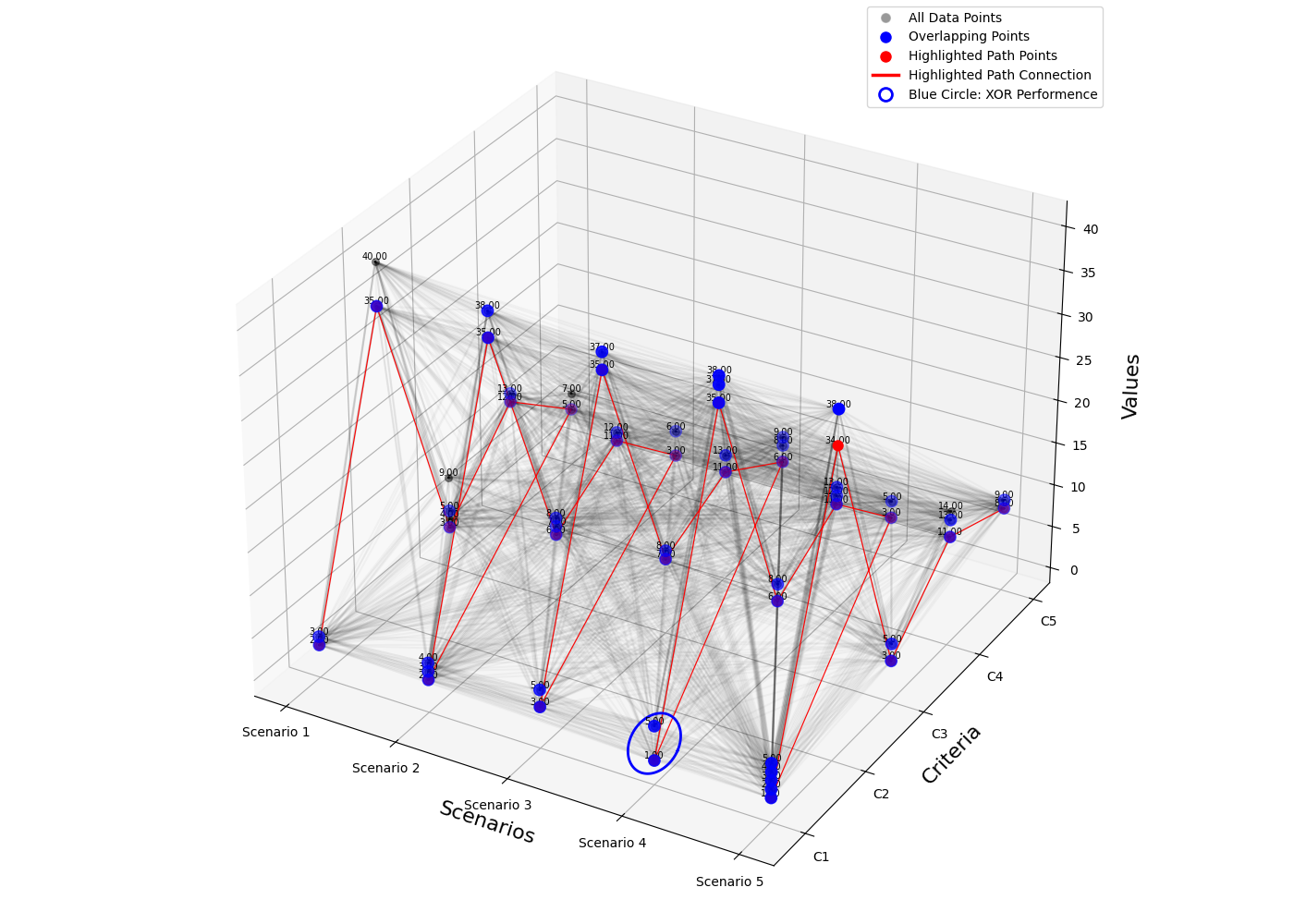}
\caption{\small Three-dimensional representation of XOR decision matrix used for cyberthreat risk assessment. The $X$, $Y$, and $Z$ axes represent the set of scenarios, set of criteria, and performance values, respectively. Each gray point (value) corresponds to a partial evaluation (i.e., element of an XOR number). The partial evaluation values collectively define the structure of XOR performance. The faint black lines connect all possible partial evaluation values within each scenario across criteria, illustrating the high-dimensional and interconnected nature of the XOR decision matrix. The blue points indicate overlapping evaluation values, where different scenarios share the same evaluation value under a given criterion. A red trajectory represents one example among potentially thousands of deterministic WSM paths embedded in the XOR matrix space.}
\label{fig:3d}
\end{figure}

\begin{table}[H]
\centering
\caption{Normalized XOR decision matrix.}
\label{tab:normalized-xor}
\resizebox{\textwidth}{!}{%
\begin{tabular}{l l l l l l}
\hline
& Execution Likelihood & Intrusion Detection Time & Severity of Grid Disruption & Response \& Mitigation Time & Attribution Difficulty \\
\hline
Scenario 1 
& $\mathsf{XOR}\{0.4000,\; 0.6000\}$ 
& $\mathsf{XOR}\{0.0000,\; 0.1250\}$ 
& $\mathsf{XOR}\{0.0000,\; 0.4444,\; 0.5556,\; 0.6667\}$ 
& $\mathsf{XOR}\{0.0000,\; 0.0769\}$ 
& $\mathsf{XOR}\{0.2222,\; 0.4444\}$ \\

Scenario 2 
& $\mathsf{XOR}\{0.2000,\; 0.4000,\; 0.6000\}$ 
& $\mathsf{XOR}\{0.0500,\; 0.1250\}$ 
& $\mathsf{XOR}\{0.1111,\; 0.2222,\; 0.3333\}$ 
& $\mathsf{XOR}\{0.0769,\; 0.1538\}$ 
& $\mathsf{XOR}\{0.3333,\; 0.6667\}$ \\

Scenario 3 
& $\mathsf{XOR}\{0.0000,\; 0.4000\}$ 
& $\mathsf{XOR}\{0.0750,\; 0.1250\}$ 
& $\mathsf{XOR}\{0.1111,\; 0.2222\}$ 
& $\mathsf{XOR}\{0.0000,\; 0.1538\}$ 
& $\mathsf{XOR}\{0.0000,\; 0.1111,\; 0.3333\}$ \\

Scenario 4 
& $\mathsf{XOR}\{0.0000,\; 0.8000\}$ 
& $\mathsf{XOR}\{0.0500,\; 0.0750,\; 0.1250\}$ 
& $\mathsf{XOR}\{0.1111,\; 0.3333\}$ 
& $\mathsf{XOR}\{0.0000,\; 0.0769,\; 0.1538\}$ 
& $\mathsf{XOR}\{0.4444,\; 0.6667\}$ \\

Scenario 5 
& $\mathsf{XOR}\{0.0000,\; 0.2000,\; 0.4000,\; 0.6000,\; 0.8000\}$ 
& $\mathsf{XOR}\{0.0500,\; 0.1500\}$ 
& $\mathsf{XOR}\{0.4444,\; 0.6667\}$ 
& $\mathsf{XOR}\{0.0000,\; 0.1538\}$ 
& $\mathsf{XOR}\{0.0000,\; 0.1111\}$ \\
\hline
\end{tabular}%
}
\end{table}

\noindent To compute the weighted normalized XOR decision matrix, we use (4.10), obtaining the results listed in Table~\ref{tab:weighted-normalized-xor}.

\begin{table}[H]
\centering
\caption{Weighted normalized XOR decision matrix.}
\label{tab:weighted-normalized-xor}
\resizebox{\textwidth}{!}{%
\begin{tabular}{l l l l l l}
\hline
& Execution Likelihood & Intrusion Detection Time & Severity of Grid Disruption & Response \& Mitigation Time & Attribution Difficulty \\
\hline
Scenario 1 
& $\mathsf{XOR}\{0.1000,\; 0.1500\}$ 
& $\mathsf{XOR}\{0.0000,\; 0.0125\}$ 
& $\mathsf{XOR}\{0.0000,\; 0.1333,\; 0.1667,\; 0.2000\}$ 
& $\mathsf{XOR}\{0.0000,\; 0.0154\}$ 
& $\mathsf{XOR}\{0.0333,\; 0.0667\}$ \\

Scenario 2 
& $\mathsf{XOR}\{0.0500,\; 0.1000,\; 0.1500\}$ 
& $\mathsf{XOR}\{0.0050,\; 0.0125\}$ 
& $\mathsf{XOR}\{0.0333,\; 0.0667,\; 0.1000\}$ 
& $\mathsf{XOR}\{0.0154,\; 0.0308\}$ 
& $\mathsf{XOR}\{0.0500,\; 0.1000\}$ \\

Scenario 3 
& $\mathsf{XOR}\{0.0000,\; 0.1000\}$ 
& $\mathsf{XOR}\{0.0075,\; 0.0125\}$ 
& $\mathsf{XOR}\{0.0333,\; 0.0667\}$ 
& $\mathsf{XOR}\{0.0000,\; 0.0308\}$ 
& $\mathsf{XOR}\{0.0000,\; 0.0167,\; 0.0500\}$ \\

Scenario 4 
& $\mathsf{XOR}\{0.0000,\; 0.2000\}$ 
& $\mathsf{XOR}\{0.0050,\; 0.0075,\; 0.0125\}$ 
& $\mathsf{XOR}\{0.0333,\; 0.1000\}$ 
& $\mathsf{XOR}\{0.0000,\; 0.0154,\; 0.0308\}$ 
& $\mathsf{XOR}\{0.0667,\; 0.1000\}$ \\

Scenario 5 
& $\mathsf{XOR}\{0.0000,\; 0.0500,\; 0.1000,\; 0.1500,\; 0.2000\}$ 
& $\mathsf{XOR}\{0.0050,\; 0.0150\}$ 
& $\mathsf{XOR}\{0.1333,\; 0.2000\}$ 
& $\mathsf{XOR}\{0.0000,\; 0.0308\}$ 
& $\mathsf{XOR}\{0.0000,\; 0.0167\}$ \\
\hline
\end{tabular}%
}
\end{table}

\begin{figure}[H]
\centering
\includegraphics[width=0.7\textwidth]{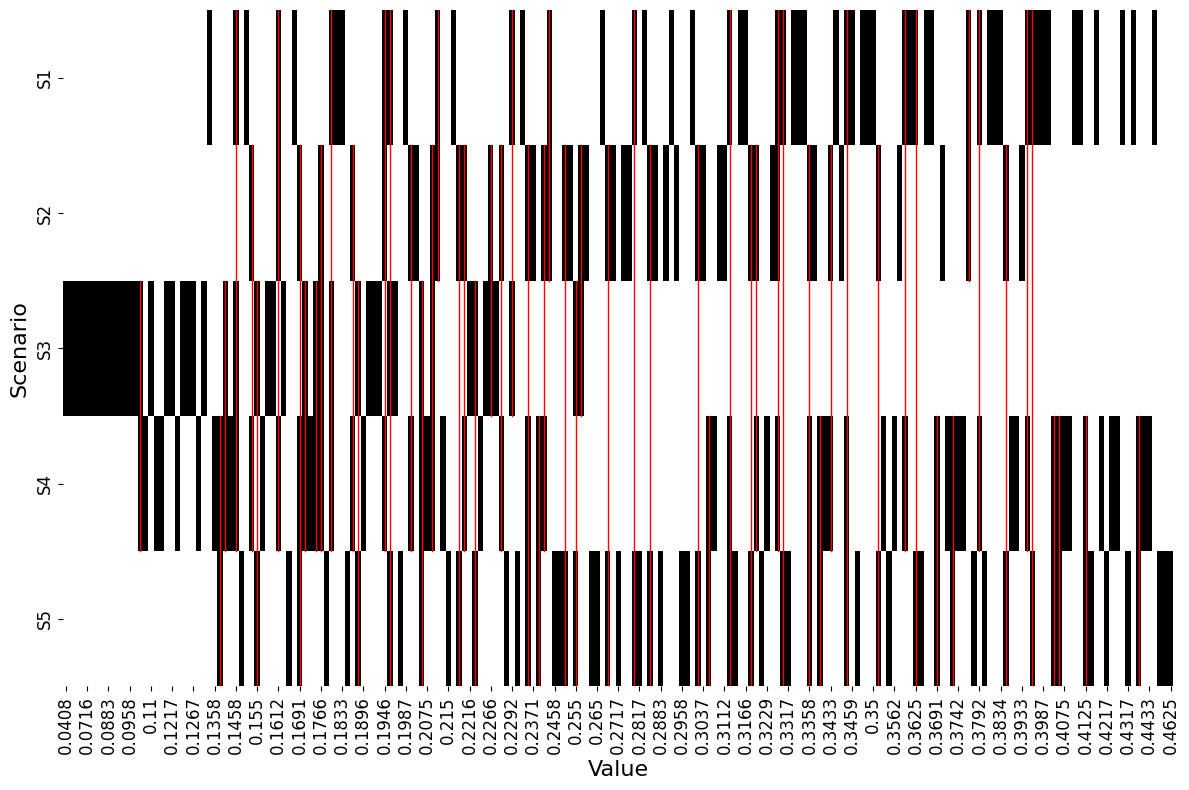}
\caption{
Overlapping global XOR-performance scores across five scenarios (S1--S5). Each cell represents either the presence (black) or absence (white) of an evaluation value. The evaluation values are shown in ascending order along the $X$ axis, while the $Y$ axis corresponds to the five scenarios. The vertical red lines indicate evaluation values shared across two or more scenarios, visually linking those with overlapping scores. This representation allows to quickly identify patterns of overlap and convergence in the global XOR-performance space.
}
\label{fig:overlap_matrix}
\end{figure}

\noindent To obtain the global XOR-performance score per alternative, we use (5.5), obtaining the results shown in Figure 4, with the complete numerical results reported in the Supplementary Materials. Figure 4 highlights the overlapping in global XOR-performance scores. It occurs when two or more scenarios share similar, partially, or fully overlapping values. Overlapping substantially complicates determining a ranking scheme using the deterministic WSM. For high overlap between global XOR performances, the XORness measure helps to distinguish alternatives.

\begin{table}[H]
\centering
\caption{XORness degrees and PXness values for cybersecurity risk assessment.}
\label{tab:xorness-degrees}
\resizebox{0.55\textwidth}{!}{%
\begin{tabular}{c c c c c c}
\hline
Scenario & $\mathsf{Perf}(\cdot)$ & $\overline{\mathsf{Perf}}(\cdot)$ & $\mathsf{Xness}(\cdot)$ & $\overline{\mathsf{Xness}}(\cdot)$ & $\mathsf{PXness}_{\gamma=0.5}(\cdot)$ \\
\hline
1 & 0.3104 & 1.0000 & 5.4638 & 0.6537 & 0.6732 \\
2 & 0.2735 & 0.8812 & 3.4655 & 0.4146 & 0.7333 \\
3 & 0.1476 & 0.4757 & 3.2451 & 0.3882 & 0.5437 \\
4 & 0.2739 & 0.8823 & 8.3585 & 1.0000 & 0.4412 \\
5 & 0.3004 & 0.9679 & 6.2572 & 0.7486 & 0.6096 \\
\hline
\end{tabular}%
}
\end{table}

For $\gamma=0.5$, applying (6.12) provides the results listed in Table~\ref{tab:xorness-degrees}. Because the criteria are oriented toward minimization, a higher $\mathsf{PXness}_{\gamma}(\cdot)$ indicates a \emph{less critical} scenario. As listed in Table~\ref{tab:xorness-degrees}, \emph{insider attacks on energy trading platforms (scenario~4)} is the most critical with the lowest $\mathsf{PXness}_{\gamma=0.5}$ ($0.4412$) among the considered scenarios. The second most critical scenario is \emph{malware injection into energy management systems (scenario~3)}, yielding a $\mathsf{PXness}_{\gamma=0.5}$ of $0.5437$. This rank is mainly driven by the low normalized performance level, $\overline {\mathsf{Perf}}(\cdot)=0.4757$. The third critical scenario is \emph{exploitation of IoT devices embedded in energy infrastructure (scenario~5)}, achieving a $\mathsf{PXness}_{\gamma=0.5}$ of $0.6096$. Despite a relatively mild underlying risk (high $\overline {\mathsf{Perf}}(\cdot)$ of $0.9679$), this scenario is associated with considerable XORness, with $\overline {\mathsf{Xness}}(\cdot)=0.7486$, which negatively influences its overall $\mathsf{PXness}_{\gamma=0.5}$. The fourth critical scenario is \emph{ransomware attacks on the regional power grid (scenario~1)}, with a $\mathsf{PXness}_{\gamma=0.5}$ of $0.6732$. Finally, \emph{coordinated cyberattacks (scenario~2)} is identified as the least critical scenario with its highest $\mathsf{PXness}_{\gamma=0.5}$ being $0.7333$.
 
As noted in Remark~8, parameter $\gamma \in [0,1]$ controls the tradeoff between performance and XORness. When $\gamma = 1$, the ranking depends solely on $\overline {\mathsf{Perf}}(\cdot)$. When $\gamma = 0$, it depends solely on $\overline{\mathsf{Xness}}(\cdot)$. Because the ranking of the cyberthreat scenarios can be sensitive to $\gamma$, the analyst might need to conduct a sensitivity analysis by varying~$\gamma$ over interval $[0,1]$ to obtain different solutions using diverse well-established sensitivity analysis techniques (see, e.g., \textcolor{blue}{Best \& Grauer, 1991}). Adapting these techniques to the XOR-WSM context is a promising direction for future applied research.

\section{Conclusion}
\noindent
Drawing on modern decision theory and information theory, we regard uncertainty as a complex and multifaceted phenomenon that is difficult, if not impossible, to capture within a single unified framework (\textcolor{blue}{Savage, 1954}; \textcolor{blue}{Bellman \& Zadeh, 1970}; \textcolor{blue}{Shafer, 1976}; \textcolor{blue}{Pawlak, 1982}; \textcolor{blue}{Deng, 1989}; \textcolor{blue}{Klir, 2006}). Attempts to either formalize all sources of uncertainty within a single framework or transform one type of uncertainty to conform to the assumptions of another often lead to oversimplification, distortion, and loss of valuable information necessary for understanding the decision problem at hand. Our aim in this study was therefore not to establish a universally superior framework, but to develop an approach suited to XORness. More broadly, we emphasize that different conditions and assumptions concerning the nature of uncertainty naturally require different formalizations and analytical tools.

\bigskip

In many practical settings, decision makers express their preferences as an exclusive alternation of the form
\begin{center}
``it is either \dots\ or \dots, but we do not know which.''
\end{center}
\noindent In such cases, uncertainty arises neither from randomness nor from vagueness but from the coexistence of several mutually exclusive and plausible evaluation values, exactly one of which is the true value, although it remains unknown which candidate is the true value. XOR numbers have been specifically developed to formalize this type of uncertainty. The notation
\[
x_1 \ \mathsf{XOR}\ x_2 \ \mathsf{XOR}\ \cdots \ \mathsf{XOR}\ x_k
\]
indicates that in a given decision context, exactly one evaluation value is realized, but its identity remains unknown. This formalism does not rely on any \emph{a priori} probabilistic hypotheses. Instead, it formalizes exclusive disjunctive alternation arising primarily from indeterminacy driven by hesitancy. Hence, XOR numbers are particularly suited to decision contexts, such as cybersecurity, extreme events, and emerging risks, where data are scarce or even unavailable. In addition, XOR-based and probabilistic modeling techniques should not be considered as incompatible. When additional information becomes available, an XOR number may be integrated into, or interpreted within, probability theory by defining a probability distribution over its elements. For example, one may externally impose a uniform distribution on the support set $\{x_1, x_2, \ldots, x_k\}$, assigning probability $1/k$ to each element. However, notably, such modeling requires additional assumptions and is not implied by the canonical form of an XOR number.

\bigskip

From this perspective, XOR- and probability-based MCDM approaches can be understood as addressing fundamentally different epistemic conditions and analytical objectives. In studies on probability-based MCDM, evaluations are typically modeled as random variables defined on a given probability space, with each evaluation being associated with a probability distribution often assumed \emph{a priori} to follow a parametric form, such as a normal or uniform distribution. Decision analysis within this framework relies on probabilistic tools such as expected values, variances, stochastic dominance, expected utility functions, and Monte Carlo simulations (\textcolor{blue}{Stewart \& Durbach, 2016}). Although this approach is well suited for situations where sufficient empirical data or reliable subjective probability assessments are available, it requires decision makers or analysts to adhere to the underlying probability assumptions. Indeed, when such assumptions are not justified, probabilistic modeling using uniform distributions may introduce information not present in the preference elicitation. In contrast, in the proposed approach, evaluations are expressed by XOR numbers without any \emph{a priori} probabilistic hypotheses. This considerably reduces the cognitive effort required during elicitation and avoids the need for probability assumptions that may be difficult to justify or validate. Operationally, XOR-based MCDM follows a different analytical pathway. It performs algebraic operations of XOR numbers for MCDM. Consequently, normalization, weighting, and aggregation can be performed with transparency and full traceability. This analysis preserves uncertainty as it is elicited with neither transformations nor adjustments that may lead to the loss of valuable information necessary for understanding the concerned decision problem. Overall, XOR-based MCDM is suitable when uncertainty arises from XORness, and probability-based MCDM is suitable when uncertainty arises from randomness.

\bigskip

Because XOR numbers represent preferences, the notion of consistency arises naturally. In deterministic preference models, preferences are often regarded as consistent when transitivity holds, and violations of transitivity are interpreted as inconsistencies. In MCDM analysis, such as the analytic hierarchy process, this notion appears in the pairwise comparison matrix, where judgments are considered consistent when they satisfy multiplicative transitivity (\textcolor{blue}{Saaty, 1980}). In real decision problems, this rule is frequently violated because of bounded rationality, incomplete information, and ill-defined decision problems (\textcolor{blue}{Ishizaka \& Nemery, 2013}). Consequently, classical transitivity may not appropriately reflect consistency under such conditions. Under XORness, the decision maker’s preferences (i.e., evaluations) are expressed by XOR numbers. When a transitivity rule is selected to assess the consistency of these preferences, one must first specify how the transitivity property is to be interpreted: internally, as examined in XOR--AHP (\textcolor{blue}{Hocine \& Kouaissah, 2020}), or externally, as explored in group XOR--AHP (\textcolor{blue}{Chen et al., 2025}). In both cases, the structure of the XOR number is consistent.\\

\noindent \textbf{Acknowledgments.} The author sincerely thanks the anonymous reviewers for their careful and professional reviews, which have positively contributed to improving the manuscript.

\vspace{1em}
\begin{center}
{\sffamily\bfseries\Large References}
\end{center}
\vspace{0.5em}

\begin{spacing}{1.1}
{\footnotesize
\begin{enumerate}[leftmargin=*, labelsep=0.5em, itemsep=2pt, parsep=0pt]

\item Ahn, B. S. (2017). The analytic hierarchy process with interval preference statements. \textit{Omega}, \textit{67}, 177--185.

\item Aissi, H., \& Roy, B. (2010). Robustness in multi-criteria decision aiding. In M. Ehrgott, J. R. Figueira, \& S. Greco (Eds.), \textit{Trends in multiple criteria decision analysis} (Vol.~142, pp.~101--139). Springer Science + Business Media Inc.

\item Arrow, K. J. (1965). \textit{Aspects of the theory of risk-bearing} (Yrj\"{o} Jahnsson Lectures). Yrj\"{o} Jahnssonin S\"{a}\"{a}ti\"{o}.

\item Baas, S. M., \& Kwakernaak, H. (1977). Rating and ranking of multiple-aspect alternatives using fuzzy sets. \textit{Automatica}, \textit{13}(1), 47--58.

\item Bana e Costa, C. A., \& Vansnick, J. C. (1994). MACBETH---An interactive path towards the construction of cardinal value functions. \textit{International Transactions in Operational Research}, \textit{1}(4), 489--500.

\item Barnes, E., \& Williams, J. R. G. (2011). A theory of metaphysical indeterminacy. In K. Bennett \& D. W. Zimmerman (Eds.), \textit{Oxford studies in metaphysics} (Vol.~6, pp.~103--148). Oxford University Press.

\item Beynon, M., Curry, B., \& Morgan, P. (2000). The Dempster--Shafer theory of evidence: An alternative approach to multicriteria decision modelling. \textit{Omega}, \textit{28}(1), 37--50.

\item Bellman, R. E., \& Zadeh, L. A. (1970). Decision-making in a fuzzy environment. \textit{Management Science}, \textit{17}(4), B-141--B-164.

\item Belton, V., \& Stewart, T. J. (2002). \textit{Multiple criteria decision analysis: An integrated approach}. Kluwer Academic Publishers.

\item Ben~Amor, S., Guitouni, A., \& Martel, J.-M. (2015). A synthesis of information imperfection representations for decision aid. \textit{INFOR: Information Systems and Operational Research}, \textit{53}(2), 68--77.

\item Best, M. J., \& Grauer, R. R. (1991). Sensitivity analysis for mean-variance portfolio problems. \textit{Management Science}, \textit{37}(8), 980--989.

\item Bhol, S. G. (2025). Applications of multi criteria decision making methods in cyber security. In A. Choudhury, K. Kaushik, V. Kumar, \& B. K. Singh (Eds.), \textit{Cyber-physical systems security: A multi-disciplinary approach} (pp.~233--258). Springer.

\item Bouyssou, D., \& Pirlot, M. (2008). On some ordinal models for decision making under uncertainty. \textit{Annals of Operations Research}, \textit{163}(1), 19--48.

\item Charnes, A., \& Cooper, W. W. (1957). Management models and industrial applications of linear programming. \textit{Management Science}, \textit{4}(1), 38--91.

\item Chen, S.-J., \& Hwang, C.-L. (1992). Fuzzy multiple attribute decision making methods. In S.-J. Chen \& C.-L. Hwang (Eds.), \textit{Fuzzy multiple attribute decision making: Methods and applications. Lecture notes in economics and mathematical systems} (Vol.~375, pp.~289--486). Springer.

\item Chen, Y. R., Liu, F., Bustince, H., Huang, H., \& Zhong, X. C. (2025). Optimization-driven group decision making with XOR comparison matrices. \textit{Mathematics and Computers in Simulation}, \textit{241}(Part-A), 763--786.

\item Corrente, S., Greco, S., Kadzi\'{n}ski, M., \& S\l{}owi\'{n}ski, R. (2013). Robust ordinal regression in preference learning and ranking. \textit{Machine Learning}, \textit{93}(2--3), 381--422.

\item Debreu, G. (1954). Representation of a preference ordering by a numerical function. In R. M. Thrall, C. H. Coombs, \& R. L. Davis (Eds.), \textit{Decision processes} (pp.~159--165). Wiley.

\item Dias, L. C. (2024). On the sigma-mu stochastic multicriteria analysis: Exact solutions for common particular cases. \textit{Omega}, \textit{127}, 103093.

\item Doumpos, M., Figueira, J. R., Greco, S., \& Zopounidis, C. (2019). \textit{New perspectives in multiple criteria decision making}. Springer International Publishing.

\item Dubois, D., \& Prade, H. (1980). \textit{Fuzzy sets and systems: Theory and applications} (Mathematics in Science and Engineering, Vol.~144). Academic Press.

\item Dyer, J. S. (2016). MAUT---Multiattribute utility theory. In S. Greco, M. Ehrgott, \& J. R. Figueira (Eds.), \textit{Multiple criteria decision analysis: State of the art surveys} (Vol.~78, pp.~285--314). Springer.

\item Dyer, J. S., \& Sarin, R. K. (1979). Measurable multiattribute value functions. \textit{Operations Research}, \textit{27}(4), 810--822.

\item Edwards, W. (1977). How to use multiattribute utility measurement for social decisionmaking. \textit{IEEE Transactions on Systems, Man, and Cybernetics}, \textit{7}(5), 326--340.

\item Fern\'{a}ndez, E., Figueira, J. R., \& Navarro, J. (2020). Interval-based extensions of two outranking methods for multi-criteria ordinal classification. \textit{Omega}, \textit{95}, 102065.

\item Figueira, J. R., Greco, S., Roy, B., \& S\l{}owi\'{n}ski, R. (2013). An overview of ELECTRE methods and their recent extensions. \textit{Journal of Multi-Criteria Decision Analysis}, \textit{20}(1--2), 61--85.

\item Fishburn, P. C. (1970). \textit{Utility theory for decision making}. Wiley.

\item Fishburn, P. C. (1978). A survey of multiattribute/multicriterion evaluation theories. In S. Zionts (Ed.), \textit{Multiple criteria problem solving. Lecture notes in economics and mathematical systems} (Vol.~155, pp.~181--224). Springer.

\item Ganin, A. A., Quach, P., Panwar, M., Collier, Z. A., Keisler, J. M., Marchese, D., \& Linkov, I. (2020). Multicriteria decision framework for cybersecurity risk assessment and management. \textit{Risk Analysis}, \textit{40}(1), 183--199.

\item Greco, S., Ehrgott, M., \& Figueira, J. R. (Eds.). (2016). \textit{Multiple criteria decision analysis: State of the art surveys}. Springer.

\item Greco, S., Matarazzo, B., \& Slowinski, R. (2001). Rough sets theory for multicriteria decision analysis. \textit{European Journal of Operational Research}, \textit{129}(1), 1--47.

\item Guitouni, A., \& Martel, J.-M. (1998). Tentative guidelines to help choosing an appropriate MCDA method. \textit{European Journal of Operational Research}, \textit{109}(2), 501--521.

\item Harper, D. (2024). \textit{Online Etymology Dictionary}. \url{https://www.etymonline.com}

\item Hocine, A., \& Kouaissah, N. (2020). XOR analytic hierarchy process and its application in the renewable energy sector. \textit{Omega}, \textit{97}, 102082.

\item Hocine, A., Kouaissah, N., \& Ortobelli~Lozza, S. (2023). XOR-analytic network process and assessing the impact of {COVID}-19 by sector. \textit{Computers \& Industrial Engineering}, \textit{177}, 109017.

\item Hosseinzadeh~Lotfi, F., Allahviranloo, T., Pedrycz, W., Shahriari, M., Sharafi, H., \& Mosavi, A. (2023). A new approach to the simple additive weighting model with triangular fuzzy numbers. In F. Hosseinzadeh~Lotfi, T. Allahviranloo, M. Shahriari, \& H. Sharafi (Eds.), \textit{Fuzzy decision analysis: Multi attribute decision making approach} (pp.~117--140). Springer.

\item Hwang, C.-L., \& Yoon, K. (1981). \textit{Multiple attribute decision making: Methods and applications. Lecture notes in economics and mathematical systems} (Vol.~186). Springer.

\item Ishizaka, A., \& Nemery, P. (2013). \textit{Multi-criteria decision analysis: Methods and software}. Wiley.

\item Jacquet-Lagr\`{e}ze, E., \& Siskos, J. (1982). Assessing a set of additive utility functions for multicriteria decision-making, the {UTA} method. \textit{European Journal of Operational Research}, \textit{10}(2), 151--164.

\item Jennings, R. E. (1994). \textit{The genealogy of disjunction}. Oxford University Press.

\item Julong, D. (1989). Introduction to grey system theory. \textit{The Journal of Grey System}, \textit{1}(1), 1--24.

\item Kahneman, D., \& Tversky, A. (1982). Variants of uncertainty. \textit{Cognition}, \textit{11}(2), 143--157.

\item Keeney, R. L., \& Raiffa, H. (1976). \textit{Decisions with multiple objectives: Preferences and value tradeoffs}. Wiley.

\item Klir, G. J. (1991). \textit{Facets of systems science}. Springer.

\item Klir, G. J. (2006). \textit{Uncertainty and information: Foundations of generalized information theory}. Wiley-IEEE Press.

\item K\"{o}ksalan, M. M., Wallenius, J., \& Zionts, S. (2011). \textit{Multiple criteria decision making: From early history to the 21st century}. World Scientific.

\item Kouaissah, N., \& Hocine, A. (2022). {XOR} data envelopment analysis and its application to renewable energy technology ranking. \textit{Expert Systems with Applications}, \textit{207}, 118044.

\item Kouaissah, N., \& Hocine, A. (2023). {XOR}-best worst method and its assessment to {COVID}-19 sectorial impact. \textit{Annals of Operations Research}, \textit{360}(1), 367--390.

\item Krantz, D. H., Luce, R. D., Suppes, P., \& Tversky, A. (1971). \textit{Foundations of measurement: Vol.~1. Additive and polynomial representations}. Academic Press.

\item Lahdelma, R., Hokkanen, J., \& Salminen, P. (1998). {SMAA}---Stochastic multiobjective acceptability analysis. \textit{European Journal of Operational Research}, \textit{106}(1), 137--143.

\item Laplace, P.-S. de. (1995). \textit{A philosophical essay on probabilities} (F. W. Truscott \& F. L. Emory, Trans.). Dover. (Original work published 1814)

\item Lindley, D. V. (1962). Indeterminate truth-values. \textit{The Journal of Philosophy}, \textit{59}(17), 449--458.

\item Liu, F., Chen, Y.-R., \& Zhou, D.-H. (2023). A two-dimensional approach to flexibility degree of {XOR} numbers with application to group decision making. \textit{Mathematics and Computers in Simulation}, \textit{207}, 267--287.

\item Liu, N., Zhang, J., Zhang, H., \& Liu, W. (2010). Security assessment for communication networks of power control systems using attack graphs and {MCDM}. \textit{IEEE Transactions on Power Delivery}, \textit{25}(3), 1492--1500.

\item Lo, C.-C., \& Chen, W.-J. (2012). A hybrid information security risk assessment procedure considering interdependences between controls. \textit{Expert Systems with Applications}, \textit{39}(1), 247--257.

\item Markowitz, H. (1952). Portfolio selection. \textit{The Journal of Finance}, \textit{7}(1), 77--91.

\item Modarres, M., \& Sadi-Nezhad, S. (2005). Fuzzy simple additive weighting method by preference ratio. \textit{Intelligent Automation \& Soft Computing}, \textit{11}(4), 235--244.

\item Moore, R. E. (1979). \textit{Methods and applications of interval analysis}. Society for Industrial and Applied Mathematics.

\item Oxford University Press. (2024). Hesitation. In \textit{Oxford English Dictionary}. \url{https://www.oed.com/}

\item Oxford University Press. (2024). Indeterminacy. In \textit{Oxford English Dictionary}. \url{https://www.oed.com/}

\item Pawlak, Z. (1982). Rough sets. \textit{International Journal of Computer \& Information Sciences}, \textit{11}(5), 341--356.

\item Quine, W. V. O. (1960). \textit{Word and object}. {MIT} Press.

\item Roy, B. (1996). \textit{Multicriteria methodology for decision aiding. Nonconvex Optimization and Its Applications} (Vol.~12). Springer.

\item Russell, B. (1962). \textit{An inquiry into meaning and truth}. Routledge.
\item Saaty, T. L. (1980). \textit{The analytic hierarchy process: Planning, priority setting, resource allocation}. {McGraw-Hill}.

\item Savage, L. J. (1954). \textit{The foundations of statistics}. Wiley.

\item Shafer, G. (1976). \textit{A mathematical theory of evidence}. Princeton University Press.

\item Shackle, G. L. S. (1969). \textit{Decision order and time in human affairs}. Cambridge University Press.

\item Stewart, T. J. (1995). Simplified approaches for multicriteria decision making under uncertainty. \textit{Journal of Multi-Criteria Decision Analysis}, \textit{4}(4), 246--258.

\item Stewart, T. J., \& Durbach, I. N. (2016). Dealing with uncertainties in {MCDA}. In S. Greco, M. Ehrgott, \& J. R. Figueira (Eds.), \textit{Multiple criteria decision analysis: State of the art surveys} (Vol.~78, pp.~467--496). Springer.

\item Triantaphyllou, E., \& Lin, C.-T. (1996). Development and evaluation of five fuzzy multiattribute decision-making methods. \textit{International Journal of Approximate Reasoning}, \textit{14}(4), 281--310.

\item Tzeng, G.-H., \& Huang, J.-J. (2011). \textit{Multiple attribute decision making: Methods and applications}. {CRC} Press.

\item Vincke, P. (1992). \textit{Multicriteria decision-aid}. John Wiley \& Sons.

\item von Neumann, J., \& Morgenstern, O. (1947). \textit{Theory of games and economic behavior}. Princeton University Press.

\item Wald, A. (1950). \textit{Statistical decision functions}. Wiley.

\item Wallenius, J., Dyer, J. S., Fishburn, P. C., Steuer, R. E., Zionts, S., \& Deb, K. (2008). Multiple criteria decision making, multiattribute utility theory: Recent accomplishments and what lies ahead. \textit{Management Science}, \textit{54}(7), 1336--1349.

\item Wang, Y.-J. (2015). A fuzzy multi-criteria decision-making model by associating technique for order preference by similarity to ideal solution with relative preference relation. \textit{Applied Soft Computing}, \textit{30}, 412--422.

\item World Economic Forum. (2025). \textit{Global cybersecurity outlook 2025}. \url{https://www.weforum.org/publications/global-cybersecurity-outlook-2025/}

\item Yager, R. R. (1988). On ordered weighted averaging aggregation operators in multicriteria decisionmaking. \textit{IEEE Transactions on Systems, Man, and Cybernetics}, \textit{18}(1), 183--190.

\item Zadeh, L. A. (1965). Fuzzy sets. \textit{Information and Control}, \textit{8}(3), 338--353.

\item Zimmermann, H.-J. (2011). \textit{Fuzzy set theory---and its applications}. Springer.

\end{enumerate}
}
\end{spacing}

\end{document}